\documentclass[conference]{IEEEtran}
\usepackage[utf8]{inputenc}

\usepackage{enumitem}
\usepackage{booktabs}
\ifCLASSINFOpdf
  \usepackage[pdftex]{graphicx}
\else
  \usepackage[dvips]{graphicx}
\fi
\graphicspath{{images/}{figures/}}
\usepackage[cmex10]{amsmath}
\usepackage{amssymb}
\usepackage[caption=false,font=footnotesize]{subfig}
\usepackage{algorithm}
\usepackage{algorithmic}
\usepackage{dblfloatfix} 

\IfFileExists{algorithmic.sty}{%
  \usepackage{algorithmic}%
}{%
  \newif\ifALCnumbered
  \newlength{\ALCindent}
  \setlength{\ALCindent}{1.5em}
  \newcounter{ALCline}
  \newcounter{ALCdepth}
  \newcommand{\ALClineprefix}{%
    \ifALCnumbered
      \makebox[2em][r]{\theALCline}\hspace*{0.5em}%
    \fi
  }
  \newcommand{\ALCnewline}{%
    \par\noindent
    \stepcounter{ALCline}%
    \ALClineprefix\hspace*{\value{ALCdepth}\ALCindent}%
  }
  \newenvironment{algorithmic}[1][0]{%
    \begingroup
    \setcounter{ALCline}{0}%
    \setcounter{ALCdepth}{0}%
    \ifnum##1>0\ALCnumberedtrue\else\ALCnumberedfalse\fi
    \parindent0pt
    \parskip0pt
  }{%
    \par\endgroup
  }
  \newcommand{\STATE}{\ALCnewline}
  \newcommand{\REQUIRE}{\ALCnewline\textbf{Inputs:} }
  
  \newcommand{\IF}[1]{\ALCnewline\textbf{if} ##1 \textbf{then}\addtocounter{ALCdepth}{1}}
  
  \newcommand{\ENDIF}{\addtocounter{ALCdepth}{-1}\ALCnewline\textbf{end if}}
  \newcommand{\WHILE}[1]{\ALCnewline\textbf{while} ##1 \textbf{do}\addtocounter{ALCdepth}{1}}
  \newcommand{\ENDWHILE}{\addtocounter{ALCdepth}{-1}\ALCnewline\textbf{end while}}
}

\usepackage{url}

\usepackage{color}
\definecolor{tumblue}{rgb}{0, 0.4, 0.74}
\usepackage{tikz}
\usetikzlibrary{arrows.meta,decorations.markings,decorations.pathreplacing,calc,positioning}

\begin{document}


%
\title{CommonRoad-Game: A Human-in-the-Loop Simulation Framework for Autonomous Driving}

\author{\IEEEauthorblockN{Yunfei Bi and Youran Wang}
\IEEEauthorblockA{Technical University of Munich\\
Munich, Germany\\
Email: \{yunfei.bi, youran.wang\}@tum.de}}


\maketitle

\begin{abstract}
Motion planning algorithms should be evaluated in
human-in-the-loop environments to ensure they produce safe and efficient behaviors during interactions. However, existing simulation platforms
often rely on recorded datasets, lack dedicated interfaces for real-time human
interaction, or remain weakly integrated with an autonomous driving ecosystem.
Moreover, many human-in-the-loop simulators are computationally intensive by design, 
making them less suitable for rapid prototyping and flexible experimentation in early-stage autonomous driving research.
To address these limitations, we present \textbf{CommonRoad-Game}, a lightweight
human-in-the-loop simulation framework tightly integrated with the CommonRoad
platform, focusing on the systematic testing of motion planners with human
participation and the analysis of human driving behaviors in interactive
scenarios. We introduce a multi-threaded architecture with a robust
synchronization mechanism that aligns simulation time with wall-clock time,
enabling deterministic and temporally consistent interaction between autonomous
and human-driven vehicles. In addition, the framework provides a scenario
generation module that records driving logs,
allowing diverse and reproducible test cases to be constructed from human-in-the-loop
experiments.
Experimental results demonstrate that \textbf{CommonRoad-Game} achieves stable
temporal synchronization, supports scalable multi-agent simulation, and
seamlessly integrates CommonRoad-compatible motion planners to generate interactive
driving scenarios.
The source code is publicly available at \url{https://github.com/Yunfei-Bi8/CommonRoad-Game}.

\end{abstract}


%
\IEEEpeerreviewmaketitle

\section{Introduction}

Making safe and efficient decisions in interactive driving scenarios is a 
critical capability for motion planning algorithms in autonomous driving. 
Driving inherently involves complex interactions between autonomous vehicles (AVs)
and human drivers, requiring the ability to understand and respond to human behavior.
In situations such as merging, yielding, or navigating shared road spaces,
where agents continuously adapt to each other's actions, even minor variations
in driving decisions may lead to substantially different outcomes~\cite{schwarting2019social}.
Consequently, reliable experimental platforms for studying human driving
behaviors and the interactions among human-driven vehicles (HVs) and AVs have become increasingly
important. However, real-world experiments with physical vehicles are often
costly, risky, and difficult to scale.

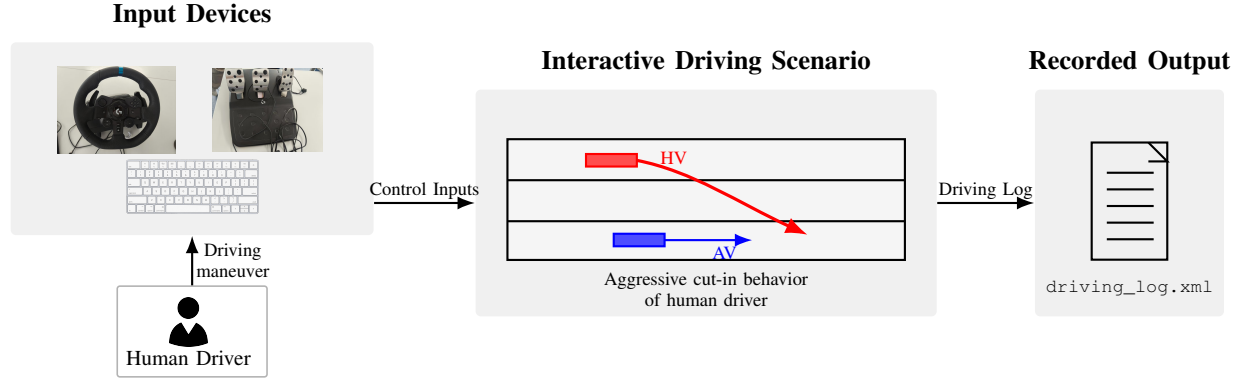
\begin{figure*}[!t]
  \centering
  \begin{tikzpicture}[
    >=Latex,
    font=\footnotesize,
    mainbox/.style={
      rectangle,
      draw=none,
      fill=gray!12,
      rounded corners=2pt,
      align=center
    },
    subbox/.style={
      rectangle,
      draw=gray!60,
      line width=0.6pt,
      fill=white,
      rounded corners=1pt,
      minimum height=0.75cm,
      align=center,
      font=\footnotesize\normalfont
    },
    title/.style={
      font=\bfseries\normalsize,
      align=center
    },
    arrow/.style={
      ->,
      thick
    },
    flowlabel/.style={
      font=\scriptsize\normalfont,
      inner sep=1.2pt,
      align=center
    },
    carhv/.style={draw=red, fill=red!70, thick},
    carav/.style={draw=blue, fill=blue!70, thick}
  ]

    \coordinate (leftcol) at (0,0);

    \node[mainbox, minimum width=4.8cm, minimum height=2.55cm]
      at ([yshift=0.85cm]leftcol) (inputdevices) {};

    \node[mainbox, minimum width=6.1cm, minimum height=3cm]
      at ([xshift=6.80cm]leftcol) (scenario) {};

    \node[mainbox, minimum width=2.5cm, minimum height=3cm]
      at ([xshift=12.40cm]leftcol) (logfile) {};

    \node[title, above=2pt of inputdevices.north] {Input Devices};
    \node[title, above=2pt of scenario.north] {Interactive Driving Scenario};
    \node[title, above=2pt of logfile.north] {Recorded Output};

    \node[inner sep=0pt]
      at ([xshift=-1.00cm,yshift=0.38cm]inputdevices.center)
      {\includegraphics[width=1.55cm]{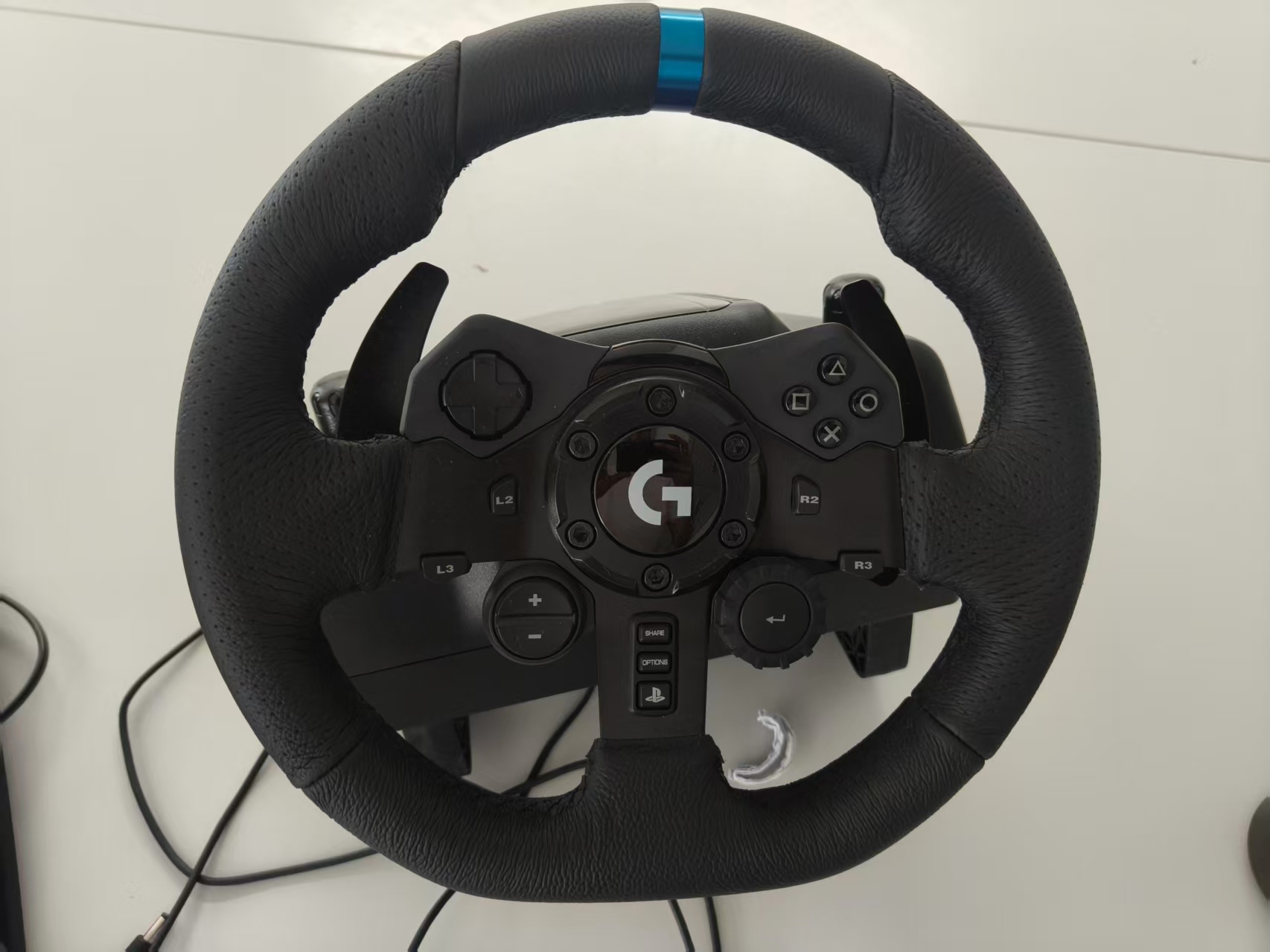}};

    \node[inner sep=0pt]
      at ([xshift=1.00cm,yshift=0.38cm]inputdevices.center)
      {\includegraphics[width=1.45cm]{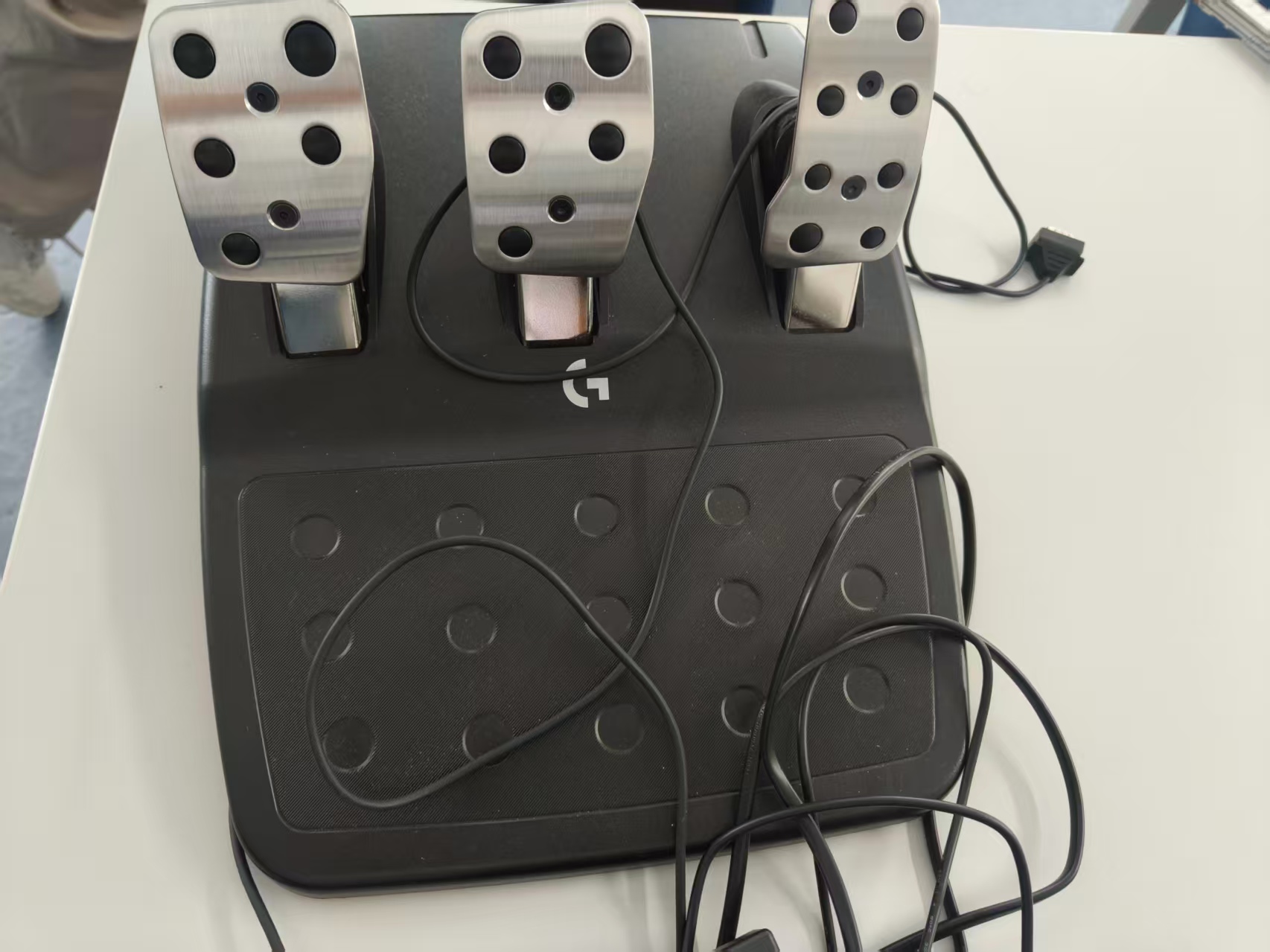}};

    \node[inner sep=0pt]
      at ([xshift=0cm,yshift=-0.65cm]inputdevices.center)
      {\includegraphics[width=1.85cm]{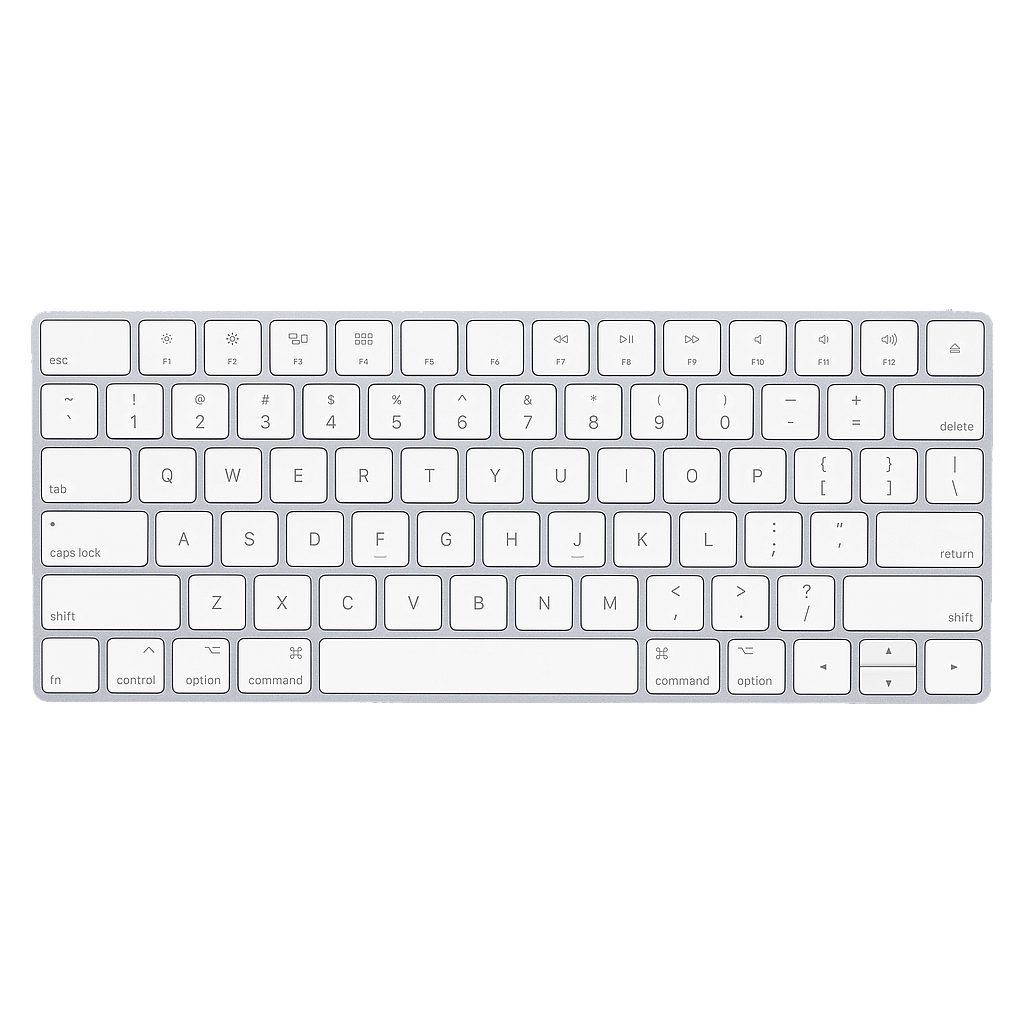}};

    \node[subbox, minimum width=1.75cm, minimum height=1.20cm]
      at ([yshift=-1.70cm]leftcol) (driver) {
        \shortstack{
          \includegraphics[height=0.65cm]{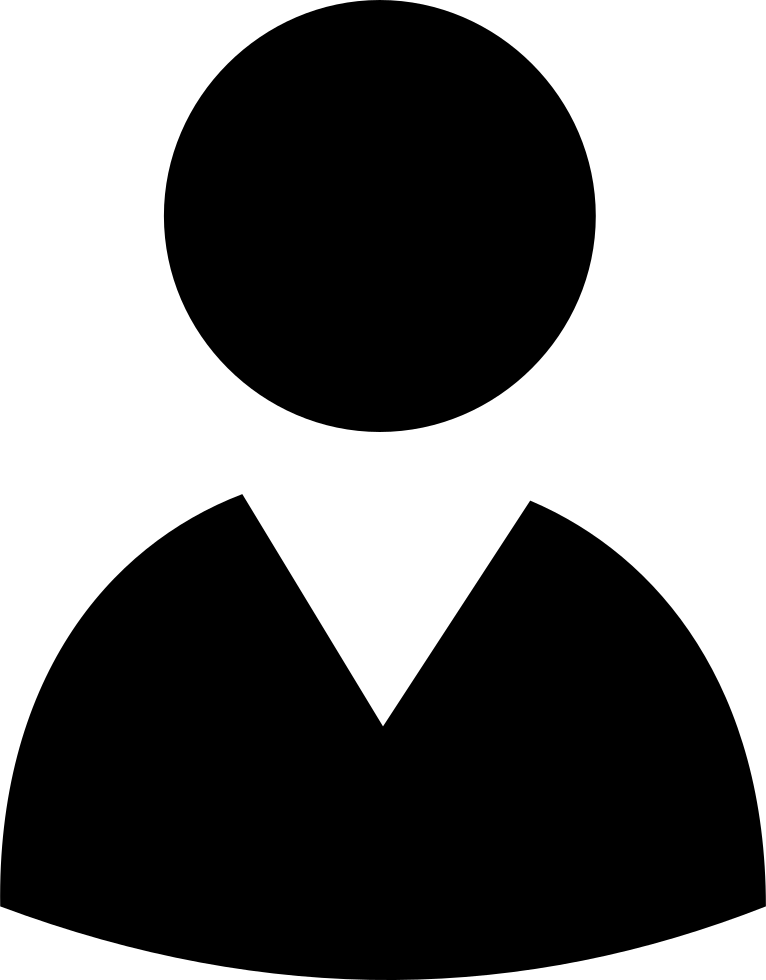}\\[-0.5mm]
          Human Driver
        }
      };

    \draw[arrow]
      (driver.north) -- (inputdevices.south)
      node[midway, right, flowlabel] {Driving\\maneuver};

    \draw[arrow]
      (inputdevices.east |- scenario.center) -- (scenario.west)
      node[midway, above, flowlabel] {Control Inputs};

    \draw[arrow]
      (scenario.east) -- (logfile.west)
      node[midway, above, flowlabel] {Driving Log};

    \begin{scope}[shift={(scenario.south west)}, x={(scenario.south east)}, y={(scenario.north west)}]
      \fill[gray!8] (0.07,0.25) rectangle (0.93,0.78);
      \draw[thick] (0.07,0.25) rectangle (0.93,0.78);

      \draw[thick] (0.07,0.42) -- (0.93,0.42);
      \draw[thick] (0.07,0.60) -- (0.93,0.60);

      \draw[carav] (0.30,0.31) rectangle ++(0.11,0.055);
      \draw[->, >=Latex, thick, blue] (0.41,0.337) -- (0.60,0.337);

      \draw[carhv] (0.24,0.66) rectangle ++(0.11,0.055);
      \draw[->, >=Latex, very thick, red]
        (0.35,0.687) .. controls (0.47,0.64) and (0.57,0.50) .. (0.72,0.36);

      \node[font=\scriptsize, text=red]  at (0.43,0.7) {HV};
      \node[font=\scriptsize, text=blue] at (0.54,0.28) {AV};

      \node[align=center, font=\scriptsize]
        at (0.50,0.12) {Aggressive cut-in behavior\\of human driver};
    \end{scope}

    \draw[thick]
      ([xshift=-0.50cm,yshift=-0.75cm]logfile.center) rectangle ++(1.00cm,1.55cm);

    \draw[thick]
      ([xshift=0.25cm,yshift=0.80cm]logfile.center) --
      ++(0.25cm,-0.25cm) --
      ++(-0.25cm,0);

    \foreach \yy in {0.40,0.18,-0.04,-0.26,-0.48} {
      \draw[thick]
        ([xshift=-0.30cm,yshift=\yy cm]logfile.center) --
        ([xshift=0.32cm,yshift=\yy cm]logfile.center);
    }

    \node[font=\ttfamily\scriptsize, align=center]
      at ([yshift=-1.12cm]logfile.center) {driving\_log.xml};

  \end{tikzpicture}
  \caption{Motivation and overall workflow of CommonRoad-Game. Human driver inputs are used to generate interactive driving behavior, including safety-critical maneuvers such as aggressive cut-ins between a human-driven vehicle (HV, red) and an autonomous vehicle (AV, blue). The resulting interaction is recorded as a structured driving log for scenario generation and analysis.}
  \label{fig:motivation_workflow}
\end{figure*}

Fig.~\ref{fig:motivation_workflow} summarizes the motivation and overall workflow
of this work: real human inputs are used to create interactive and potentially
critical driving behaviors, which are then recorded as structured driving logs
for subsequent scenario generation and analysis.

To address these challenges, we present \textbf{CommonRoad-Game}, a lightweight
human-in-the-loop simulation framework that enables interactive evaluation of
motion planners within the CommonRoad~\cite{althoff2017commonroad}\footnote{CommonRoad website:
\url{https://commonroad.in.tum.de/}} ecosystem and facilitates the transition
from offline benchmarking toward more realistic human-in-the-loop testing.

In the experimental evaluation, we demonstrate the applicability of the framework
with two representative motion planning algorithms: an Intelligent Driver Model
(IDM)-based motion planner\footnote{\label{fn:idm_planner}CommonRoad IDM planner: \url{https://gitlab.lrz.de/cps/commonroad/commonroad-idm-planner}}
and a reactive sampling-based motion planner\footnote{\label{fn:reactive_planner}Reactive planner: \url{https://gitlab.lrz.de/cps/reactive-planner}}.
Both planners are available within the CommonRoad ecosystem.

\subsection{Related Work}
This section reviews representative simulation platforms for autonomous driving
research, with a particular focus on simulation fidelity, human-in-the-loop
capability, and support for motion planning evaluation.

High-fidelity, general-purpose simulators such as CARLA~\cite{dosovitskiy2017carla},
LGSVL Simulator~\cite{rong2020lgsvl}, and AirSim~\cite{shah2017airsim}
provide realistic 3D environments with configurable sensors and physics.
CARLA~\cite{dosovitskiy2017carla} is widely used for developing, training, and
validating autonomous driving systems under diverse sensor suites and
environmental conditions.
LGSVL~\cite{rong2020lgsvl} emphasizes integration with full autonomy stacks such
as Autoware\footnote{Autoware: Open-Source Software Stack for Autonomous Driving, \url{https://www.autoware.org/}} and Apollo\footnote{Apollo: An Open Autonomous Driving Platform, \url{https://www.apollo.auto/}}, supporting
both Software-in-the-Loop (SIL) and Hardware-in-the-Loop (HIL) testing.
AirSim~\cite{shah2017airsim}, built on Unreal Engine, supports both aerial and
ground vehicles and enables software- and hardware-in-the-loop experiments via
standard communication interfaces (e.g., MAVLink~\cite{MAVLinkGuide}).

Beyond high-fidelity simulators, MetaDrive~\cite{li2023metadrive} focuses on
scalable scenario generation for reinforcement learning, enabling diverse traffic
configurations via procedural composition and real-traffic data replay.
Its design primarily targets large-scale data generation and policy training,
rather than controlled interactive studies with human participants.

Closer to lightweight human-in-the-loop settings, CARLO~\cite{cao2020reinforcement}\footnote{CARLO repository: \url{https://github.com/Stanford-ILIAD/CARLO}}
(\textit{CARLA-Low Budget}) is a 2D simulator designed for HVs.
It provides an interface to consumer-grade driving hardware, a simplified vehicle
model for mapping real-time control inputs to trajectories, and a lightweight
visualization environment.
CARLO has proven effective for studying human driving behavior in controlled
settings, particularly for reinforcement learning research.

Overall, existing simulation platforms reflect different design trade-offs.
High-fidelity simulators prioritize visual realism, sensor modeling, and
full-stack autonomy, often at the cost of increased system complexity and
computational overhead.
Lightweight simulators enable rapid prototyping and human-in-the-loop studies,
yet typically provide limited support for structured motion planning evaluation
under standardized benchmark formats.
These observations motivate simulation frameworks that balance lightweight design
with systematic support for interactive planner evaluation.

\subsection{Contributions}
Inspired by CARLO~\cite{cao2020reinforcement}, we develop \textbf{CommonRoad-Game},
a lightweight human-in-the-loop simulation framework tightly integrated with the
CommonRoad~\cite{althoff2017commonroad} ecosystem. Compared with CARLO, which
focuses on HV simulation, and with general-purpose platforms
such as MetaDrive~\cite{li2023metadrive}, AirSim~\cite{shah2017airsim}, LGSVL
Simulator~\cite{rong2020lgsvl}, and CARLA~\cite{dosovitskiy2017carla},
CommonRoad-Game is designed for systematic interactive planner evaluation while
remaining lightweight. The main contributions are summarized as follows:
\begin{itemize}
    \item \textbf{Interactive Evaluation of Motion Planners:} We provide a
    standardized setup for assessing the safety and robustness of CommonRoad
    motion planners in scenarios involving real human participants.
    \item \textbf{Traffic Scenario Generation:} We support recording and replay
    of human driving logs, with direct export to the CommonRoad
    scenario format,
    enabling the collected interaction data to be reused for training
    data-driven driving models as well as for systematic evaluation and
    benchmarking of motion planners.
    \item \textbf{Behavioral Capture and Analysis:} We provide high-fidelity
    logging of human driving behavior to support quantitative analysis of
    human--AV interactions.
    \item \textbf{Multi-Agent and Multi-Interface Simulation:} We support
    multi-AV scenarios and multiple human input interfaces
    (e.g., keyboard and steering wheels) for HVs, while
    maintaining temporal consistency between simulator execution and planner
    scenario time through a multi-threaded architecture.
\end{itemize}

\section{Preliminaries}
\subsection{CommonRoad}
CommonRoad~\cite{althoff2017commonroad}\footnote{CommonRoad website:
\url{https://commonroad.in.tum.de/}} is a benchmark framework and scenario
format for motion planning of road vehicles. It aims to make motion-planning
experiments reproducible and comparable by describing the relevant elements of a
planning problem in a standardized form, including the road network, static and
dynamic obstacles, vehicle models, planning goals, constraints, and cost
functions. In this way, a benchmark can be specified by a compact identifier
while still providing the information needed to reconstruct the corresponding
planning problem.

In this work, CommonRoad~\cite{althoff2017commonroad} serves as the common representation layer between
human-in-the-loop simulation and motion planner execution. HV
states recorded in the simulator are mapped to CommonRoad-compatible dynamic
obstacles, while planner outputs are interpreted as trajectories in the
CommonRoad scenario frame. This standardized representation enables the proposed
framework to reuse existing CommonRoad scenarios and planners, and to export
recorded interactions as CommonRoad scenario files for offline analysis and
benchmarking.

\subsection{Curvilinear Coordinate Frame}
Since our framework requires AVs and HVs to be tightly synchronized
in one simulation frame (the global Cartesian frame), 
while the motion planners sourced from CommonRoad~\cite{althoff2017commonroad} employed in this work usually perform planning in a curvilinear (Frenet) 
coordinate frame along a reference path, coordinate transformations between Cartesian and curvilinear 
coordinates are required. In our implementation, Frenet-coordinate conversion is
performed using the robust curvilinear coordinate transformation toolbox of
W\"{u}rsching and Althoff~\cite{wuersching2024robust}.

A curvilinear coordinate frame is defined with respect to a given reference path $\Gamma$~\cite{wuersching2024robust}.
Instead of representing a planar position by Cartesian coordinates $\mathbf{p}=(x,y)^{\mathrm{T}}$,
the curvilinear representation uses (i) the arc-length $s$ measured along $\Gamma$ and
(ii) the lateral (orthogonal) deviation $d$ from the path point $\Gamma(s)$.
Accordingly, a Cartesian point is mapped to curvilinear coordinates through a transformation
\begin{equation}
T_{\Gamma}: \mathbb{R}^2 \to \mathbb{R}^2,\quad (x,y) \mapsto (s,d).
\label{eq:cart2curv}
\end{equation}

\subsection{Vehicle Model}
In this work, we model the vehicle motion in the global Cartesian frame using the kinematic bicycle (single-track) model~\cite{althoff2017commonroad}.

\begin{figure}[t]
  \centering
  \begin{tikzpicture}[scale=0.85, line cap=round, line join=round, >=Latex]

    \draw[->, line width=0.9pt] (0,0) -- (8.2,0) node[below right] {$x$};
    \draw[->, line width=0.9pt] (0,0) -- (0,6.2) node[above left] {$y$};

    \coordinate (R) at (2.2,1.6);   
    \coordinate (F) at (5.3,4.7);   

    \def\deltadeg{35} 

    \pgfmathsetmacro{\alpha}{atan2(4.7-1.6,5.3-2.2)}

    \draw[line width=1.2pt] (R) -- (F);

    \begin{scope}[shift={(R)}, rotate=\alpha]
      \draw[line width=1.0pt] (0,0) ellipse (0.65 and 0.28);
      \fill (0,0) circle (0.04);
    \end{scope}

    \begin{scope}[shift={(F)}, rotate=\alpha+\deltadeg]
      \draw[line width=1.0pt] (0,0) ellipse (0.75 and 0.28);
      \fill (0,0) circle (0.04);
    \end{scope}

    \draw[line width=0.6pt, gray!60] (R) -- ($(R)+(\alpha-90:0.55)$);
    \node[right, xshift=0pt] at ($(R)+(\alpha-90:0.55)$) {$(x,y)$};

    \draw[->, line width=0.9pt, gray!70]
      ($(R)!0.38!(F)$) -- ($(R)!0.58!(F)$)
      node[midway, above left, black] {$v$};

    \draw[line width=0.6pt, gray!60] (R) -- ($(R)+(-0.55,0.55)$);
    \draw[line width=0.6pt, gray!60] (F) -- ($(F)+(-0.55,0.55)$);
    \draw[<->, line width=0.8pt, gray!60]
      ($(R)+(-0.55,0.55)$) -- ($(F)+(-0.55,0.55)$)
      node[midway, left=2pt, black] {$L$};

    \draw[line width=0.6pt] (F) -- ($(F)+(2.0,0)$);        
    \draw[line width=0.6pt] (F) -- ($(F)+(\alpha:1.8)$);   
    \draw[->, line width=0.8pt]
      ($(F)+(0:1.2)$) arc[start angle=0, end angle=\alpha, radius=1.2]
      node[pos=0.65, right] {$\theta$};

    \draw[line width=0.6pt] (F) -- ($(F)+(\alpha:1.3)$);                 
    \draw[line width=0.6pt] (F) -- ($(F)+(\alpha+\deltadeg:1.3)$);     

    \draw[->, line width=0.8pt]
      ($(F)+(\alpha:0.95)$)
      arc[start angle=\alpha, end angle=\alpha+\deltadeg, radius=0.95]
      node[pos=0.55, above, xshift=3pt] {$\delta$};

  \end{tikzpicture}
  \caption{Kinematic bicycle (single-track) model in the global Cartesian frame.}
  \label{fig:kinematic_bicycle_model}
\end{figure}
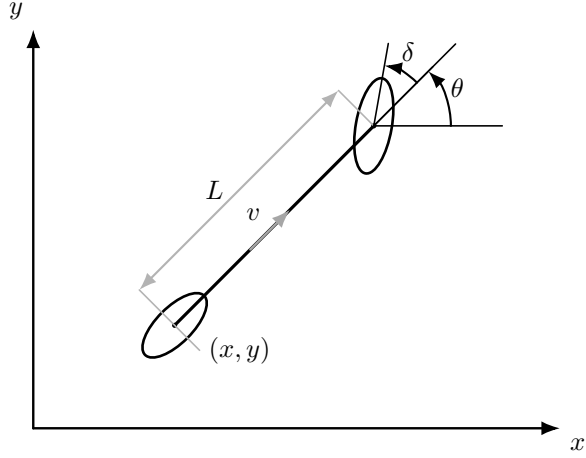

The state-space representation of kinematic bicycle model is given by
\begin{equation}
\dot{\mathbf{x}} = f(\mathbf{x}, \mathbf{u}),
\label{eq:state_space}
\end{equation}
where the state vector is
\begin{equation}
\mathbf{x} = [x, y, \theta, v, \delta]^{\mathrm{T}},
\label{eq:state_vector}
\end{equation}
and the input vector is
\begin{equation}
\mathbf{u} = [a, \dot{\delta}]^{\mathrm{T}},
\label{eq:input_vector}
\end{equation}
where the state variables are: $(x,y) \in \mathbb{R}^2$ denoting the rear-axle position in the global Cartesian frame, $\theta$ the heading angle, $v$ the longitudinal velocity along the vehicle heading direction, and $\delta$ the front-wheel steering angle.
The input variables are: $a$ the longitudinal acceleration and $\dot{\delta}$ the steering rate.
The parameter $L$ denotes the wheelbase of the vehicle.

\section{Methodology}

\begin{figure*}[!t]
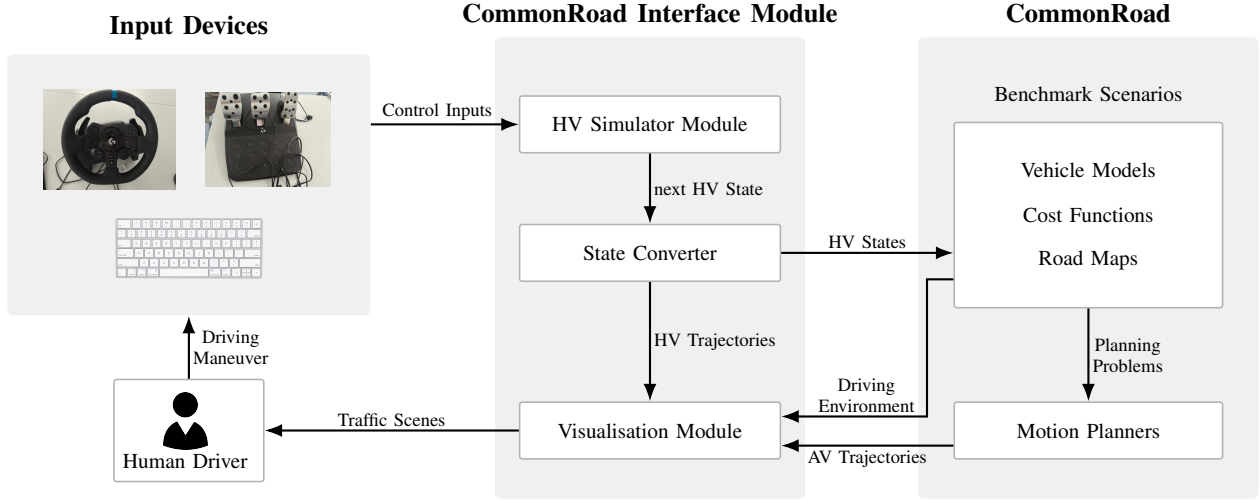

  \centering
  \begin{tikzpicture}[
    >=Latex,
    mainbox/.style={
      rectangle,
      draw=none,
      fill=gray!12,
      rounded corners=2pt,
      align=center
    },
    subbox/.style={
      rectangle,
      draw=gray!60,
      line width=0.6pt,
      fill=white,
      rounded corners=1pt,
      minimum height=0.75cm,
      align=center,
      font=\footnotesize\normalfont
    },
    title/.style={
      font=\bfseries\normalsize,
      align=center
    },
    arrow/.style={
      ->,
      thick
    },
    flowlabel/.style={
      font=\scriptsize\normalfont,
      inner sep=1.2pt,
      align=center
    }
  ]

    \coordinate (leftcol) at (0,0);

    \node[mainbox, minimum width=4.8cm, minimum height=3.45cm]
      at ([yshift=1.1cm]leftcol) (inputdevices) {};

    \node[mainbox, minimum width=4.1cm, minimum height=6.1cm]
      at ([xshift=6.10cm]leftcol) (interface) {};

    \node[mainbox, minimum width=4.5cm, minimum height=6.1cm]
      at ([xshift=11.90cm]leftcol) (commonroad) {};

    \node[title, above=2pt of inputdevices.north] {Input Devices};
    \node[title, above=2pt of interface.north] {CommonRoad Interface Module};
    \node[title, above=2pt of commonroad.north] {CommonRoad};

    \node[inner sep=0pt]
      at ([xshift=-1.05cm,yshift=0.60cm]inputdevices.center)
      {\includegraphics[width=1.75cm]{figures/steering-wheel.png}};

    \node[inner sep=0pt]
      at ([xshift=1.05cm,yshift=0.60cm]inputdevices.center)
      {\includegraphics[width=1.65cm]{figures/pedal.png}};

    \node[inner sep=0pt]
      at ([xshift=0cm,yshift=-0.85cm]inputdevices.center)
      {\includegraphics[width=2.05cm]{figures/keyboard.png}};

    \node[subbox, minimum width=1.75cm, minimum height=1.35cm]
      at ([yshift=-2.15cm]leftcol) (driver) {
        \shortstack{
          \includegraphics[height=0.75cm]{figures/humandriver.png}\\[-0.5mm]
          Human Driver
        }
      };

    \draw[arrow]
      (driver.north) -- (inputdevices.south)
      node[midway, right, flowlabel] {Driving\\Maneuver};

    \node[subbox, minimum width=3.45cm]
      at ([yshift=1.90cm]interface.center) (hvsim) {HV Simulator Module};

    \node[subbox, minimum width=3.45cm]
      at ([yshift=0.20cm]interface.center) (converter) {State Converter};

    \node[subbox, minimum width=3.45cm]
      at ([yshift=-2.15cm]interface.center) (viz) {Visualisation Module};

    \node[font=\footnotesize\normalfont, align=center]
      at ([yshift=2.3cm]commonroad.center) {Benchmark Scenarios};

    \node[subbox, minimum width=3.55cm, minimum height=2.45cm]
      at ([yshift=0.70cm]commonroad.center) (crdata) {};

    \node[font=\footnotesize\normalfont]
      at ([yshift=0.60cm]crdata.center) {Vehicle Models};

    \node[font=\footnotesize\normalfont]
      at ([yshift=0.00cm]crdata.center) {Cost Functions};

    \node[font=\footnotesize\normalfont] (roadmaps)
      at ([yshift=-0.60cm]crdata.center) {Road Maps};

    \node[subbox, minimum width=3.55cm]
      at ([yshift=-2.15cm]commonroad.center) (planner) {Motion Planners};

    \draw[arrow]
      (inputdevices.east |- hvsim.west) -- (hvsim.west)
      node[xshift=-0.1cm, midway, above, flowlabel] {Control Inputs};

    \draw[arrow]
      (hvsim.south) -- (converter.north)
      node[midway, right, flowlabel] {next HV State};

    \draw[arrow]
      (converter.south) -- (viz.north)
      node[midway, right, flowlabel] {HV Trajectories};

    \draw[arrow]
      (viz.west) -- (driver.east)
      node[midway, above, flowlabel] {Traffic Scenes};

    \draw[arrow]
      (converter.east) -- (crdata.west |- converter.east)
      node[midway, above, flowlabel] {HV States};

    \draw[arrow]
      (crdata.south) -- (planner.north)
      node[midway, right, flowlabel] {Planning\\Problems};

    \draw[arrow]
      ([yshift=-0.20cm]planner.west) -- ([yshift=-0.20cm]viz.east)
      node[midway, below, flowlabel] {AV Trajectories};

    \coordinate (mapout) at ([yshift=-0.85cm]crdata.west);
    \draw[arrow]
      (mapout) -- ++(-0.35cm,0) |- ([yshift=-0.20cm]viz.north east)
      node[midway, above, xshift=-0.8cm, flowlabel] {Driving\\Environment};;

  \end{tikzpicture}
  \caption{Overview of the CommonRoad-Game simulation framework architecture.}
  \label{fig:system_architecture}
\end{figure*}

We first include a drivability detection module that monitors the HV
footprint for collisions and road-boundary violations using the CommonRoad~\cite{althoff2017commonroad}
drivability checker package, then describe the remaining system components.
The CommonRoad-Game simulation framework mainly consists of the following components:
\begin{enumerate}[label=\arabic*), noitemsep, topsep=0pt]
    \item a \textit{CommonRoad Interface Module} that bridges the HV Simulator Module and motion planners by converting the simulation states
    of HVs into planner-compatible representations,
    and by mapping the resulting trajectories generated by motion planners back to the simulation domain.
    Through this state conversion and trajectory mapping process, the module ensures coordinate-frame and time synchronization between HV and AV behaviors,

    \item a \textit{Multi-threaded Simulation Architecture} that decouples the HV Simulator Module and Visualisation Module
    from the computationally intensive replanning process, enabling their execution in separate threads
    to ensure and analyze the synchronization between simulation time and real-world time,

    \item an \textit{HV Simulator Module} that acquires real-time control inputs from a Logitech G923 steering wheel and pedals\footnote{Logitech G923 product page: \url{https://www.logitechg.com/de-de/shop/p/g923-trueforce-sim-racing-wheel}}and outputs the trajectory of HVs according to the control inputs,

    \item a \textit{Drivability Detection Module} that checks collision and road-compliance
    of the HV using the CommonRoad drivability-checker package\footnote{\url{https://commonroad.in.tum.de/tools/drivability-checker}},

    \item a \textit{Visualisation Module} that translates CommonRoad scenarios into a customized visualization environment and visualizes the states of both AVs and HVs
    in a real-time manner,

    \item a \textit{Scenario Generation Module} that records the states of both AVs and HVs during simulations,
    converts these states into CommonRoad scenarios, and visualizes the scenarios in a frame-wise manner.
\end{enumerate}
An overview of these components is illustrated in Fig.~\ref{fig:system_architecture}.

\subsection{CommonRoad Interface Module}
The CommonRoad Interface Module synchronizes HVs and AVs along two dimensions.
First, it enforces \emph{spatial synchronization} by establishing a fixed rigid-body transform between the simulator-side and the planner-side anchors of the global metric frame and applying it whenever states or trajectories are exchanged.
Second, it enforces \emph{temporal synchronization} by updating the HV and
AV states in lockstep within each simulation cycle.

The following paragraphs detail these two aspects, respectively.

\subsubsection{Spatial Synchronization}

The simulator and the motion planner share the same global metric coordinate system defined by the CommonRoad~\cite{althoff2017commonroad} scenario, within which the HV and the AV are tracked and the planning problem is specified. At initialization, however, the simulator places the AV at a pose computed from the lanelet geometry, which is in general not identical to the initial AV pose declared in the planning problem. Spatial synchronization compensates for this initial-pose discrepancy through a fixed rigid-body transformation and two dedicated submodules that consistently map states and trajectories between the simulator-side and planner-side references throughout the simulation. For brevity, these two references are subsequently denoted as the \emph{simulation coordinate system} and the \emph{scenario coordinate system}, respectively; both correspond to the same underlying CommonRoad metric frame, but are anchored at the simulator-side and the planning-problem-declared initial AV poses.

\paragraph{Low-Level Simulation Module}
The Low-Level Simulation Module advances the states of the HV and the AV at every simulation step by numerically integrating the kinematic bicycle model. Given the road network and initial vehicle poses provided by a CommonRoad~\cite{althoff2017commonroad} scenario, the module receives the driver's real-time control inputs (steering and acceleration/brake) and the trajectory states produced by the motion planner, and outputs the updated HV state $x_h^{\mathrm{sim}}$ and AV state $x_a^{\mathrm{sim}}$ at each simulation step. Both states are expressed in a shared simulation coordinate system:
\begin{equation}
\begin{aligned}
  x_h^{\mathrm{sim}} &= [x_h, y_h, \delta_h, v_h, \psi_h]^\top, \\
  x_a^{\mathrm{sim}} &= [x_a, y_a, \delta_a, v_a, \psi_a]^\top.
\end{aligned}
\end{equation}

\paragraph{Frame-Alignment Module}
To absorb the initial-pose discrepancy between the simulation and scenario coordinate systems, the Frame-Alignment Module estimates a rigid-body transform $T_{s\rightarrow c} = (R,\mathbf{d},\Delta\psi)$ from the simulation coordinate system ($s$) to the scenario coordinate system ($c$), together with its inverse $T_{c\rightarrow s}$. The transform is computed once at initialization, held fixed throughout the simulation, and used to consistently map poses between the two references.

Denote the simulator-side and scenario-side initial AV poses by $(p_a^0, \psi_a^0)$ and $(\bar{p}_a^0, \bar{\psi}_a^0)$, respectively, where $p_a^0, \bar{p}_a^0 \in \mathbb{R}^2$ are planar positions, $\psi_a^0, \bar{\psi}_a^0$ are yaw angles, and the superscript $0$ indicates the initial time step. Let $\mathrm{wrap}_\pi(\cdot)$ denote the operator that normalizes an angle to $(-\pi,\pi]$, ensuring angular consistency and avoiding discontinuities at $\pm\pi$.

The module computes the yaw offset
\begin{equation}
  \Delta\psi = \mathrm{wrap}_\pi\!\left(\bar{\psi}_a^0 - \psi_a^0\right),
\end{equation}
the rotation matrix
\begin{equation}
  R = \begin{bmatrix}
        \cos\Delta\psi & -\sin\Delta\psi \\
        \sin\Delta\psi &  \cos\Delta\psi
      \end{bmatrix},
\end{equation}
and the translation vector
\begin{equation}
  \mathbf{d} = \bar{p}_a^0 - R\, p_a^0.
\end{equation}

For any pose $(p^{\mathrm{sim}}, \psi^{\mathrm{sim}})$ in the simulation coordinate system, the
corresponding pose $(p^{\mathrm{sc}}, \psi^{\mathrm{sc}})$ in the scenario coordinate system is given
by
\begin{equation}
\begin{aligned}
  p^{\mathrm{sc}}   &= R\, p^{\mathrm{sim}} + \mathbf{d}, \\
  \psi^{\mathrm{sc}} &= \mathrm{wrap}_\pi\!\left( \psi^{\mathrm{sim}} + \Delta\psi \right),
\end{aligned}
\end{equation}
and conversely
\begin{equation}
\begin{aligned}
  p^{\mathrm{sim}}   &= R^\top (p^{\mathrm{sc}} - \mathbf{d}), \\
  \psi^{\mathrm{sim}} &= \mathrm{wrap}_\pi\!\left( \psi^{\mathrm{sc}} - \Delta\psi \right).
\end{aligned}
\end{equation}






\paragraph{Human-Vehicle Scenario Adapter}

The HV is presented to the planner as a dynamic obstacle through an adapter module. The adapter takes as input the current human state $x_h^{\mathrm{sim}}$ in the simulation coordinate system, the rigid transform $T_{s\rightarrow c}$ between the simulation and scenario coordinate systems, the current discrete scenario time index $k$, the simulation step size $\Delta t$, and a prediction horizon $H_{\mathrm{pred}}$. It produces a dynamic obstacle in the scenario coordinate system whose current state matches the simulated HV at time index $k$, together with a short-horizon predicted trajectory of the same obstacle.

The adapter maps the simulated pose of the HV from the
simulation coordinate system to the scenario coordinate system using
$T_{s\rightarrow c}$, and updates the corresponding dynamic obstacle state
at time step $k$ with the resulting position, yaw angle, and velocity.
It then generates a short-horizon prediction over $H_{\mathrm{pred}}$ by
forward propagating a constant-velocity, constant-steering kinematic model,
expressed entirely in the scenario coordinate system.
This provides the planner with a temporally and spatially consistent
representation of the HV as a dynamic obstacle.

\paragraph{Autonomous Vehicle Interface}
Distinct from the Frame-Alignment Module, which computes the transform $T_{s\rightarrow c}$ once at initialization, this module \emph{applies} the (now fixed) transform bidirectionally at runtime. At every replanning trigger (Algorithm~\ref{alg:sync_timeline}, lines~8--10), the current AV state $x_{\mathrm{AV}}$ is mapped to the scenario frame via $T_{s\rightarrow c}$ and packaged---together with velocity, yaw, acceleration, steering angle, yaw rate, and the scenario time index $k$---into the planner's expected initial-condition format $x_{\mathrm{AV}}^{\mathrm{CR}}$. At every simulation step (Algorithm~\ref{alg:sync_timeline}, line~13), the state $\tau[\lfloor \sigma \rfloor]$ retrieved from the current planner trajectory is mapped back via $T_{c\rightarrow s}$ and overwrites the AV simulation state. Because planner conventions differ (the reactive planner expects rear-axle positions, whereas the IDM planner uses center positions), the position is conditionally shifted between the rear axle and the vehicle center using the known wheelbase in the corresponding direction.

\subsubsection{Time Synchronization}
\label{subsubsec:time_sync}

The framework involves two distinct time discretizations that, without explicit coordination, would drift apart and produce inconsistent state updates. The simulator advances at a nominal step $\Delta t$ used to integrate the HV and AV dynamics, while the motion planner and the CommonRoad~\cite{althoff2017commonroad} scenario both operate on a coarser scenario time grid with step $\Delta t_c$ that determines the temporal spacing of dynamic obstacle trajectories, predictions, and the planner's trajectory states. The two steps are in general unequal: $\Delta t$ is chosen by the simulator to be fine-grained so as to maintain real-time responsiveness to human input, whereas $\Delta t_c$ is fixed by the scenario description and reflects the logical resolution at which the scenario evolves. Furthermore, the wall-clock cost of each simulation step is itself variable due to rendering, drivability checks, and I/O overhead, and the replanning routine is computationally expensive with non-constant runtime.

If left uncoordinated, these two cadences induce three concrete failure modes. First, the simulator's wall-clock cadence diverges from $\Delta t$ because per-step computation and replanning routinely stall the main loop, breaking the real-time responsiveness required for human-in-the-loop interaction. Second, the AV's executed motion drifts off the planner's intended trajectory because the trajectory is sampled at incorrect progress indices when $\Delta t$ does not match $\Delta t_c$. Third, replanning requests fall on simulation steps that do not align with scenario-time boundaries, so the planner is re-initialized at a state inconsistent with the scenario's discrete time grid.

\begin{figure*}[!t]
\centering
\footnotesize
\begin{tikzpicture}[>=Latex, line cap=round, line join=round]
  \def\step{3.5}
  \def\yh{1.5}
  \def\ya{-1.2}
  \def\yk{-3.0}

  \foreach \i in {0,1,2,3} {
    \draw[densely dashed, gray!60] (\i*\step,\yh) -- (\i*\step,\ya);
  }
  \foreach \i in {0,3} {
    \draw[densely dashed, gray!60] (\i*\step,\ya) -- (\i*\step,\yk);
  }

  \draw[->, line width=0.6pt] (0,0) -- (3*\step+1.0,0) node[right] {simulation time};

  \foreach \i/\lab in {0/$n$,1/$n\!+\!1$,2/$n\!+\!2$,3/$n\!+\!3$} {
    \draw (\i*\step,0.06) -- (\i*\step,-0.06);
    \node[below, fill=white, inner sep=1.5pt] at (\i*\step,-0.06) {\lab};
  }

  \draw[gray!50] (0,\yh) -- (3*\step,\yh);
  \draw[gray!50] (0,\ya) -- (3*\step,\ya);
  \draw[gray!50] (0,\yk) -- (3*\step,\yk);

  \node[left] at (0,\yh) {HV update};
  \node[left] at (0,\ya) {AV update};
  \node[left] at (0,\yk) {Scenario time};

  \foreach \i in {0,1,2,3} {
    \fill (\i*\step,\yh) circle (1.2pt);
    \fill (\i*\step,\ya) circle (1.2pt);
  }

  \foreach \i/\lab in {0/$k$,3/$k\!+\!1$} {
    \draw[fill=black] (\i*\step-0.06,\yk-0.06) rectangle (\i*\step+0.06,\yk+0.06);
    \node[below] at (\i*\step,\yk-0.12) {\lab};
  }

  \draw[<->] (0,\yh+0.4) -- (\step,\yh+0.4) node[midway, above] {$\Delta t_{\mathrm{wc}}$};
  \draw[<->] (0,\yk-0.6) -- (3*\step,\yk-0.6) node[midway, below] {$N$ steps $\approx \Delta t_c$};

  \draw[->] (0.25,\yh-0.05) -- (0.25,\yk+0.08)
    node[pos=0.25, right, align=left, fill=white, inner sep=1pt] {sync HV\\replan request};
  \draw[->] (3*\step-0.15,\yh-0.05) -- (3*\step-0.15,\yk+0.08);
  \draw[->] (0,\ya-0.6) -- (3*\step,\ya-0.6) node[midway, below] {sample trajectory, update $\sigma$};

  \begin{scope}[shift={(0,-4.7)}, font=\scriptsize]
    \fill (0,0) circle (1.4pt);
    \node[anchor=west] at (0.2,0) {HV/AV state update};
    \draw[fill=black] (4.0-0.07,-0.07) rectangle (4.0+0.07,0.07);
    \node[anchor=west] at (4.2,0) {Scenario time index ($k$)};
    \draw[densely dashed, gray!60] (8.5,-0.15) -- (8.5,0.15);
    \node[anchor=west] at (8.65,0) {Inter-timeline connector};
  \end{scope}
\end{tikzpicture}
\caption{Aligned HV/AV updates within the simulation loop. Each simulation step advances both vehicles by the measured wall-clock duration $\Delta t_{\mathrm{wc}}$, while the scenario time index $k$ advances every $N$ steps (approximately one $\Delta t_c$), triggering obstacle synchronization and replanning. The example uses $N=3$.}
\label{fig:sync_timeline_diagram}
\end{figure*}
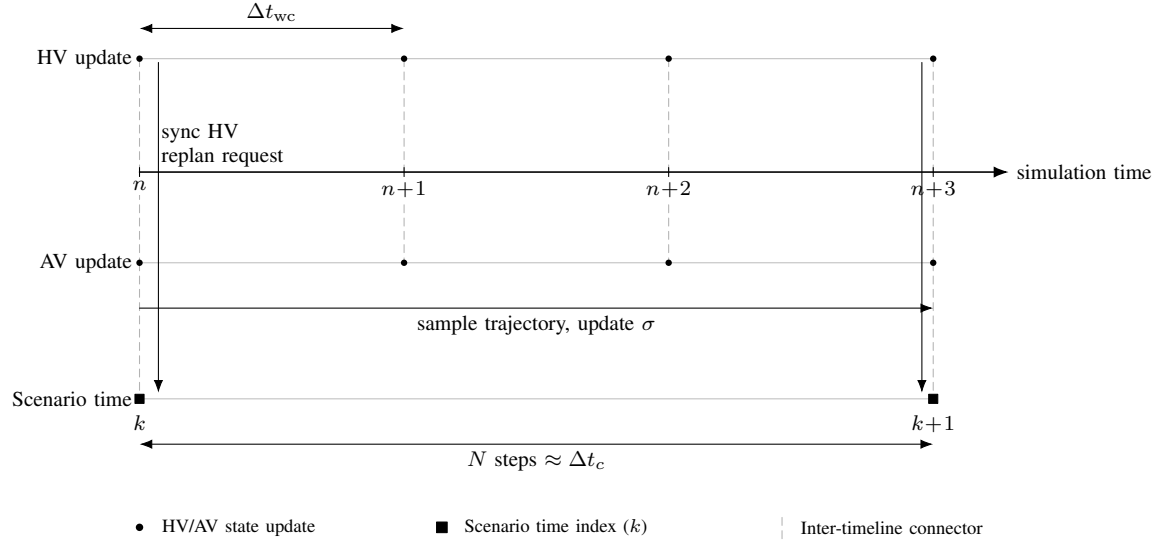

To prevent these drifts, temporal synchronization updates HV and AV states in lockstep on a shared simulation timeline, while the planner maintains its internal scenario time indexing. This synchronization is realized through a set of submodules that coordinate state updates across the simulation timeline and the planner's scenario time base.
Algorithm~\ref{alg:sync_timeline} presents a high-level pseudocode description of the
time-synchronization procedure executed in each simulation cycle.
Complementarily, Fig.~\ref{fig:sync_timeline_diagram} visualizes the aligned HV/AV updates and the corresponding scenario-step boundaries within one simulation loop.


\begin{algorithm}[t]
\caption{Synchronization of HV and AV Timeline}
\label{alg:sync_timeline}
\footnotesize
\begin{algorithmic}[1]
\REQUIRE $\Delta t$, $\Delta t_c$, $T_{s\rightarrow c}$, $k_0$
\STATE $N \leftarrow \max\!\left(1,\, \mathrm{round}(\Delta t_c / \Delta t)\right)$
\STATE $n \leftarrow 0$,\;$k \leftarrow k_0$,\;$\sigma \leftarrow 0$,\;$\textit{stop} \leftarrow \textbf{false}$,\;$\Delta t_{\mathrm{wc}} \leftarrow \Delta t$,\;$t_0 \leftarrow$ wall-clock
\WHILE{$\neg\, \textit{stop}$}
    \IF{new planner result $\tau_{\mathrm{new}}$ available}
        \STATE $\tau \leftarrow \tau_{\mathrm{new}}$,\;\; $\sigma \leftarrow \arg\min_{i}\, \big\|T_{c\rightarrow s}(\tau[i]) - x_{\mathrm{AV}}\big\|$
    \ENDIF
    \STATE $u_h \leftarrow \textsc{ReadInput}()$,\;\; $x_h \leftarrow \Phi_{\mathrm{KS}}(x_h,\, u_h,\, \Delta t_{\mathrm{wc}})$
    \IF{$n \bmod N = 0$}
        \STATE $o_h(k) \leftarrow T_{s\rightarrow c}(x_h)$
        \STATE $x_{\mathrm{AV}}^{\mathrm{CR}} \leftarrow T_{s\rightarrow c}(x_{\mathrm{AV}})$,\;\; $\textsc{EnqueueReplan}(x_{\mathrm{AV}}^{\mathrm{CR}},\, k)$
        \STATE $k \leftarrow k + 1$
    \ENDIF
    \STATE $x_{\mathrm{AV}} \leftarrow T_{c\rightarrow s}\!\left(\tau[\lfloor \sigma \rfloor]\right)$
    \STATE $\sigma \leftarrow \sigma + \Delta t / \Delta t_c$
    \IF{$\textsc{Collision}(x_h)\,\vee\,\textsc{OffRoad}(x_h)\,\vee\,\textsc{UserStop}$}
        \STATE $\textit{stop} \leftarrow \textbf{true}$
    \ENDIF
    \STATE $\textsc{WaitUntil}\!\left(t_0 + (n+1)\Delta t\right)$,\;\; $\Delta t_{\mathrm{wc}} \leftarrow$ measured interval,\;\; $n \leftarrow n + 1$
\ENDWHILE
\end{algorithmic}
\end{algorithm}

\noindent\textbf{Notation.}
The framework involves \emph{two distinct time discretizations}, both expressed in seconds. Their relationships are visualized in Fig.~\ref{fig:sync_timeline_diagram}, whose three horizontal rows correspond to the HV update cadence, the AV update cadence, and the discrete scenario timeline, respectively.

\emph{(i) Nominal simulation step $\Delta t$} is the simulator's target tick interval at which the HV and AV states are advanced, indexed by the integer simulation step index $n$. It is a design choice of the simulator and is set fine-grained (typically $10\,\mathrm{ms}$ in our implementation) to maintain real-time responsiveness to human input.

\emph{(ii) Scenario step $\Delta t_c$} is a scalar constant denoting the time interval (in seconds) between two consecutive points on the scenario timeline. It is declared by the CommonRoad~\cite{althoff2017commonroad} scenario file, is used to represent time-varying scenario elements (dynamic obstacle trajectories, predictions, and signal states), and is in general coarser than $\Delta t$ (typically $100\,\mathrm{ms}$ in our implementation). The motion planner adopts the same step: its output trajectory is a sequence of states sampled at intervals of $\Delta t_c$.

The associated \emph{scenario time index} $k$, distinct from $\Delta t_c$, is a dimensionless integer counter identifying the current position along this scenario timeline, related to $\Delta t_c$ through the identity that scenario time equals $k\,\Delta t_c$ seconds. The index is initialized at $k_0$ (declared by the planning problem) and is incremented by one every $N = \mathrm{round}(\Delta t_c / \Delta t)$ simulation steps, i.e., once per scenario interval; Fig.~\ref{fig:sync_timeline_diagram} illustrates the case $N=3$, where the filled squares mark consecutive indices $k$ and $k\!+\!1$ and their spacing on the timeline equals $\Delta t_c$. The continuous progress variable $\sigma$ further tracks the AV's sub-step sampling position along the planner trajectory between two consecutive scenario indices.

The two steps are unequal because they are fixed by different components: $\Delta t$ is chosen by the simulator for fine HV/AV control granularity, whereas $\Delta t_c$ is chosen by the CommonRoad scenario for logical-time resolution of the scenario and the planner. The time-synchronization mechanism aligns these two grids at scenario boundaries while keeping the per-step HV/AV update fine-grained.

The nominal step $\Delta t$ is to be distinguished from the \emph{realized} wall-clock interval $\Delta t_{\mathrm{wc}}$, which is the actual elapsed real time between two consecutive cycles as measured by the global time-controller. Since no measured interval is yet available at the first cycle, $\Delta t_{\mathrm{wc}}$ is initialized to the nominal step $\Delta t$ (Algorithm~\ref{alg:sync_timeline}, line~2). Under the real-time invariant maintained by the framework, $\Delta t_{\mathrm{wc}}\!\approx\!\Delta t$, but the two may transiently diverge under heavy computational load.

The remaining symbols are: $u_h$ and $x_h$ denote the human control input and the HV simulation state, respectively; $x_{\mathrm{AV}}$ is the AV simulation state, and $x_{\mathrm{AV}}^{\mathrm{CR}}$ is its mapped state in the scenario coordinate system via $T_{s\rightarrow c}$. The planner trajectory is denoted by $\tau$, a discrete sequence of states whose $i$-th element is written as $\tau[i]$, and the function $\Phi_{\mathrm{KS}}(x,u,\Delta t)$ denotes the one-step propagation of the kinematic single-track model from state $x$ under control input $u$ over the duration $\Delta t$. The HV dynamic-obstacle representation written into the scenario at index $k$ is denoted by $o_h(k)$, and $t_0$ is the wall-clock time recorded at simulation start. The stop flag is raised either by user termination or by drivability violations (collision or off-road).

We now describe the individual submodules
that together realize the proposed time-synchronization mechanism.
\paragraph{Global Time-Controller Module}
This module enforces the real-time invariant that the simulation's virtual clock---a discrete counter advancing by $\Delta t$ per control cycle---ticks at the same rate as the continuous wall-clock time reported by the operating system. Without enforcement, the per-cycle compute load would let the two timelines drift apart; the controller absorbs this drift by anchoring every cycle to a scheduled wall-clock deadline. The corresponding operations are highlighted in Algorithm~\ref{alg:sync_timeline}.

This invariant is non-trivial to maintain because the per-cycle compute load is inherently variable. The motion planner, for instance, can take substantially longer than $\Delta t$ to produce a new trajectory under heavy traffic or short scenario-time intervals, in which case it would fail to deliver a fresh trajectory in time for the next replanning boundary; even when its execution is decoupled onto a worker thread (Section~\ref{subsubsec:decouple_replan}), the main loop still incurs per-step costs for rendering, drivability checking, and the I/O acquisition of human control inputs, all of which are non-constant and can transiently exceed the step budget. Left uncompensated, these fluctuations would (i)~let the virtual clock fall behind wall-clock time, breaking the real-time responsiveness required for HITL interaction with the human driver, and (ii)~cause the HV dynamics to be integrated over the nominal $\Delta t$ rather than the actual elapsed interval, leading to physically inconsistent motion. The controller therefore needs to both pace the main loop against a wall-clock schedule and report the realized step duration to downstream consumers.

\begin{figure}[t]
\centering
\footnotesize
\begin{tikzpicture}[>=Latex, line cap=round, line join=round]
  \def\step{1.6}
  \def\ywc{1.2}
  \def\ysim{0}

  \draw[->, line width=0.6pt] (0,\ysim) -- (5.6,\ysim) node[right] {simulation time $n\Delta t$};
  \draw[->, line width=0.6pt] (0,\ywc) -- (5.6,\ywc) node[right] {wall-clock time $\tau$};

  \foreach \i/\lab in {0/$n$,1/$n\!+\!1$,2/$n\!+\!2$,3/$n\!+\!3$} {
    \draw (\i*\step,\ysim+0.05) -- (\i*\step,\ysim-0.05);
    \node[below] at (\i*\step,\ysim-0.05) {\lab};
  }

  \draw[<->] (0,\ysim-0.55) -- (\step,\ysim-0.55) node[midway, below] {$\Delta t$};

  \foreach \i in {0,1,2,3} {
    \draw (\i*\step,\ywc) circle (1.3pt);
    \draw[densely dotted, gray!60] (\i*\step,\ywc) -- (\i*\step,\ysim);
  }

  \foreach \i/\shift in {0/0.0,1/0.35,2/0.6,3/0.4} {
    \fill (\i*\step+\shift,\ywc) circle (1.2pt);
  }

  \draw[decorate, decoration={brace, amplitude=3pt, raise=3pt}, gray!70]
    (\step,\ywc) -- (\step+0.35,\ywc)
    node[midway, above=5pt, black] {$\tau_{n+1}-\tau_{n+1}^\star$};

  \draw[<->] (\step+0.35,\ywc-0.35) -- (2*\step+0.6,\ywc-0.35)
    node[midway, below] {$\tau_{n+2}-\tau_{n+1}$};

  \node[align=left, anchor=west] at (3.2,\ywc+0.6) {\scriptsize open: ideal $\tau_n^\star$\\\scriptsize filled: actual $\tau_n$};
\end{tikzpicture}
\caption{Wall-clock and simulation time alignment. Open markers on the wall-clock axis indicate the \emph{ideal} cycle-start schedule $\tau_n^\star = t_0 + n\Delta t$, while the bottom axis advances uniformly in simulation steps $n\Delta t$. Filled markers indicate the \emph{actual} wall-clock time $\tau_n$ at which each cycle begins; $\tau_n$ lags $\tau_n^\star$ whenever the per-cycle computation exceeds the step budget $\Delta t$.}
\label{fig:time_sync_timeline}
\end{figure}
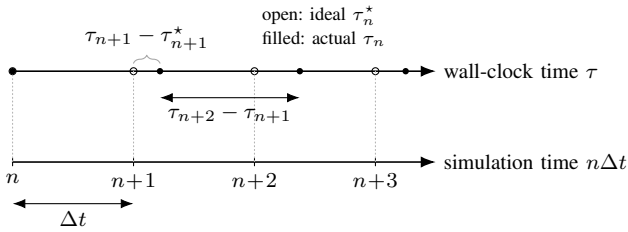

At simulation start, the controller records the wall-clock anchor $t_0$ (Algorithm~\ref{alg:sync_timeline}, line~2). Each subsequent simulation cycle~$n$ is then assigned an \emph{ideal} (scheduled) wall-clock start time
\begin{equation}
\tau_n^\star = t_0 + n\,\Delta t,
\end{equation}
i.e., the wall-clock instant at which cycle $n$ is \emph{supposed} to begin. Let $\tau_n$ denote the corresponding \emph{actual} wall-clock time at which cycle $n$ does begin. At the end of every cycle, the controller blocks the main loop in $\textsc{WaitUntil}(\tau_{n+1}^\star)$ until the scheduled deadline of cycle~$n{+}1$ is reached (Algorithm~\ref{alg:sync_timeline}, line~18); upon resuming, it measures the actual elapsed wall-clock interval since the start of the just-finished cycle, $\Delta t_{\mathrm{wc}} = \tau_{n+1} - \tau_n$, exposes the result downstream, and increments~$n$. Fig.~\ref{fig:time_sync_timeline} illustrates the resulting alignment between the ideal $\tau_n^\star$, the realized $\tau_n$, and the uniform simulation steps $n\Delta t$.

To quantify the deviation between realized execution and the ideal schedule, the controller tracks the per-step timing error and its accumulation,
\begin{equation}
\begin{aligned}
e_n &= \Delta t_{\mathrm{wc}} - \Delta t, \\
E_n &= \sum_{i=1}^{n} e_i,
\end{aligned}
\end{equation}
and exposes both quantities together with the per-cycle sleep duration for upstream monitoring. Three complementary mechanisms then maintain the alignment under variable computational load:
\begin{enumerate}
    \item \textbf{Wall-clock pacing.} The blocking wait described above enforces $\tau_n \geq \tau_n^\star$ whenever computation completes early. If the per-cycle computation exceeds the available budget, the wait reduces to zero and the simulation temporarily lags behind the ideal schedule.

    \item \textbf{Adaptive frame skipping.} When the synchronization lag exceeds a tolerance threshold, the rendering pipeline is temporarily suppressed; bypassing visualization significantly reduces per-step compute, allowing the loop to run faster than real time and gradually amortize the accumulated drift \emph{without} modifying the reference time.

    \item \textbf{Drift reset.} As a last-resort fallback, the controller re-anchors the reference time $t_0$ to the current wall-clock and resets $E_n$ whenever $|E_n|$ exceeds a preset threshold, preventing unbounded long-term drift. The reset is suppressed during a brief warm-up window at the start of simulation (the first few simulation steps) so that one-off startup transients---first-frame rendering, scenario loading, JIT compilation---do not artificially inflate $E_n$ and trigger spurious resets.
\end{enumerate}

Downstream, the controller clips the measured $\Delta t_{\mathrm{wc}}$ to a configured range $[\Delta t_{\min}, \Delta t_{\max}]$ to obtain the time step that is actually used to numerically integrate the HV kinematic single-track dynamics,
\begin{equation}
\Delta t_{\mathrm{eff}} \leftarrow \mathrm{clip}(\Delta t_{\mathrm{wc}},\, \Delta t_{\min},\, \Delta t_{\max}),
\end{equation}
and the HV state is then propagated by $\Phi_{\mathrm{KS}}(x_h, u_h, \Delta t_{\mathrm{eff}})$ at the top of the next cycle (Algorithm~\ref{alg:sync_timeline}, line~7; for brevity we write this argument as $\Delta t_{\mathrm{wc}}$ in Algorithm~\ref{alg:sync_timeline}). The integrated HV motion thus reflects the actual elapsed real time while remaining numerically stable under transient overruns. The combination of wall-clock-anchored scheduling, drift-handling mechanisms, and measured-interval integration enforces the per-cycle invariant $\Delta t_{\mathrm{wc}}\!\approx\!\Delta t$ under normal load.

\paragraph{Scenario-Time and Replanning Scheduler}

Since the planner operates on its own time step $\Delta t_c$, a dedicated scheduler links the simulation time step $\Delta t$ and the scenario time step $\Delta t_c$ by computing an integer replanning period $N$ in simulation steps and emitting a scenario time index $k$ that advances every $N$ simulation steps, thereby aligning the simulation step index $n$ with the scenario time index $k$. Here, $\Delta t$ is supplied by the global time-controller module, $\Delta t_c$ is obtained from the scenario description, and the initial scenario time index $k_0$ is specified by the planning problem.

The replanning period is chosen as
\begin{equation}
  N = \max\!\left(1,\ \mathrm{round}\!\left(
        \frac{\Delta t_c}{\Delta t}
      \right)\right),
\end{equation}
so that $N$ simulation steps correspond approximately to one scenario
time step. A scenario index $k$ is initialized at $k_0$ and updated as
follows:
\begin{itemize}
  \item The simulation step index $n$ increments by one every cycle.
  \item When $n$ is a multiple of $N$, the system forms a new planning
        problem using the current AV (and human-obstacle) state at
        scenario time $k$ and then increments $k \leftarrow k+1$.
\end{itemize}

This ensures that, roughly every $\Delta t_c$ seconds of simulated
time, the scenario time index advances by one, and the planner is
re-initialized at the correct scenario time.

\paragraph{Trajectory-Progress Tracker}

Between replanning events, the AV executes the most recent planner trajectory while accounting for the different time discretizations used by the simulator and the planner. The planner trajectory $\tau$ is represented as a discrete sequence of states sampled on the scenario time grid with step size $\Delta t_c$. At each simulation step, the tracker maintains a continuous trajectory-progress variable $\sigma$, interpreted as a continuous index into this discrete sequence, and emits the planner state at progress $\sigma$ to drive the AV update. The dynamics of $\sigma$ within one simulation loop are visualized at the bottom of Fig.~\ref{fig:sync_timeline_diagram} by the arrow labeled \emph{``sample trajectory, update $\sigma$''}: at every simulation step on the AV row, $\sigma$ is incremented and the trajectory is resampled, with $\sigma$ traversing the scenario timeline between two consecutive scenario indices $k$ and $k\!+\!1$.

The progress variable is advanced according to
\begin{equation}
  \sigma \leftarrow \sigma + \frac{\Delta t}{\Delta t_c},
\end{equation}
so that a unit increase in $\sigma$ corresponds to one scenario time step (Algorithm~\ref{alg:sync_timeline}, line~14). The current AV state is then obtained by sampling the trajectory at the discrete index $\lfloor \sigma \rfloor$ and mapping the sampled state back to the simulation coordinate system via the inverse transform $T_{c\rightarrow s}$ (Algorithm~\ref{alg:sync_timeline}, line~13).

When a new trajectory $\tau_{\mathrm{new}}$ arrives from the planner, $\tau$ is replaced by $\tau_{\mathrm{new}}$ and $\sigma$ is realigned to the discrete state whose mapped simulation pose is closest to the current AV pose,
\begin{equation}
  \sigma \leftarrow \arg\min_{i}\, \big\|T_{c\rightarrow s}(\tau[i]) - x_{\mathrm{AV}}\big\|,
\end{equation}
so as to avoid discontinuities in the executed motion (Algorithm~\ref{alg:sync_timeline}, line~5).


\subsection{Multi-threaded Simulation Architecture}
\label{subsec:mt_realtime_arch}

We employ a simulation main loop that is intentionally 
decoupled from the computationally intensive replanning process by running replanning 
in a dedicated worker thread. 
The design has two objectives: (i) to maintain a 
main loop at a target update rate, and 
(ii) to update the state of AV
whenever a new trajectory from the planner becomes available, 
without stalling the per-step simulation progression.
Here, the \emph{main loop} refers to the per-step simulation control loop running on the
main thread (state updates, trajectory application, rendering, and logging), and the
\emph{target update rate} is the nominal step period $\Delta t$
maintained by the wall-clock time controller.

\subsubsection{Decoupling the Replanning Workload from the Main Loop}
\label{subsubsec:decouple_replan}

This section details how the decoupling is realized at the thread level. We first introduce the concurrency primitives that couple the main thread and the worker thread, then describe in turn the per-step pipeline executed on the main thread, the conditions under which a replanning request is issued, the worker-thread routine that produces a new trajectory, and the way each completed result is applied back to the main loop.

The simulation framework maintains the asynchronous replanning mechanism
using a mutex and a condition variable
along
with a single-slot request/result buffer.

At each simulation step $n$, the main thread executes the following sequence:
\begin{enumerate}
    \item \textbf{Update of the current AV trajectory.} 
    If a newly computed planner trajectory is available from the worker thread, the shared \emph{current-trajectory} buffer is updated; otherwise, the previously stored trajectory is retained.
    
    \item \textbf{Propagation of the HV.}
    The current human control input is applied to the low-level vehicle model in the \emph{simulation module} to advance the HV by one step and update its visualization pose.
    The corresponding CommonRoad~\cite{althoff2017commonroad} representation (dynamic obstacle in the scenario frame) is \emph{not} updated at every step; it is refreshed only when a replanning request is assembled via obstacle synchronization.

    \item \textbf{Advancement of the AV by trajectory tracking.}
    The AV state at this simulation step is obtained deterministically from the \emph{currently stored} planner trajectory $\tau$, which is a discrete sequence of states $\tau[0], \tau[1], \ldots$ defined in the scenario frame, by retrieving the state at index $\lfloor \sigma \rfloor$. The resulting state is then mapped back to the simulation frame via the inverse transform $T_{c\rightarrow s}$ and used to overwrite the AV pose and velocity for this step, \emph{without} invoking the planner. This is the same operation as in Algorithm~\ref{alg:sync_timeline}, line~13.
    
    \item \textbf{Non-blocking periodic replanning trigger.} 
    Every $N$ simulation steps, a replanning request is formed and enqueued such that typically $N\Delta t \approx \Delta t_c$. 
    
\end{enumerate}


The full workflow is summarized in Algorithm~\ref{alg:thread_exchange}, where the main thread runs at every simulation step (lines~1--10) and the worker thread runs independently (lines~11--17); the two threads cooperate through a single request slot \texttt{req}, a single result slot \texttt{res}, and a boolean busy flag \texttt{b} (\texttt{true} when the worker is processing a request), with $N$ denoting the replanning interval in simulation steps and $n$ the current simulation step index. The remainder of this paragraph walks through this workflow from the main thread's perspective.

\emph{Triggering a new replan.} At every simulation step, the main thread evaluates the joint condition $n \bmod N = 0 \,\wedge\, \texttt{b} = \mathbf{false}$ before issuing a replan: the first conjunct enforces the scheduled interval and the second ensures the worker is idle (Algorithm~\ref{alg:thread_exchange}, line~6). When both hold, the main thread packages a request containing the current AV state mapped to the scenario frame via $T_{s\rightarrow c}$ together with the synchronized HV obstacle, hands the request to the worker, and continues without waiting (Algorithm~\ref{alg:thread_exchange}, lines~7--8). When either conjunct fails, the trigger is skipped, so the main loop never blocks on an in-progress planning task.

\emph{Worker execution.} The worker thread blocks until a request arrives (Algorithm~\ref{alg:thread_exchange}, line~12), runs the motion planner---a computation whose runtime may vary and substantially exceed $\Delta t$---and delivers the resulting trajectory back through the result slot before marking itself idle (Algorithm~\ref{alg:thread_exchange}, lines~16--17). Because the planner runs on its own thread, its variable runtime does not constrain the per-step cadence of the main loop.

\emph{Applying a delivered result.} At the start of every subsequent simulation step, the main thread first checks for a newly delivered trajectory; if one is present, it adopts the new trajectory and realigns the progress variable $\sigma$ to the discrete state whose mapped simulation pose is closest to the current AV pose (Algorithm~\ref{alg:thread_exchange}, lines~2--3; this realignment is the same operation as in Algorithm~\ref{alg:sync_timeline}, line~5). Between issuing a request and receiving its result, the AV continues to track the most recent trajectory; when a fresher one arrives it is incorporated without any waiting, which is how the loop achieves full decoupling from the planner's variable runtime.

\begin{algorithm}[t]
\caption{Asynchronous replanning workflow and shared request/result exchange between the main simulation loop and the replanning worker}
\label{alg:thread_exchange}
\footnotesize
\begin{algorithmic}[1]
\REQUIRE Replanning period $N$, step index $n$, shared request slot \texttt{req}, shared result slot \texttt{res}, busy flag \texttt{b}, mutex and condition variable

\STATE \textbf{Main thread loop (executed at each simulation step)}
\IF{\texttt{res} is available}
    \STATE Apply trajectory; reset progress; clear \texttt{res}
\ENDIF
\STATE Update HV state; retrieve AV state from the latest trajectory
\IF{$n \bmod N = 0$ \AND \texttt{b} $=$ \textbf{false}}
    \STATE Package replanning inputs (AV state in scenario frame, synchronized obstacles)
    \STATE Write \texttt{req}; set \texttt{b} $\leftarrow$ \textbf{true}; notify worker
\ENDIF
\STATE Advance to next step without waiting for replanning

\STATE \textbf{Worker thread loop}
\STATE Wait on condition until \texttt{req} is available or shutdown is requested
\IF{shutdown is requested}
    \STATE Exit worker loop
\ENDIF
\STATE Copy \texttt{req}; clear request slot
\STATE Run planner; write \texttt{res}; set \texttt{b} $\leftarrow$ \textbf{false}; notify main thread
\end{algorithmic}
\end{algorithm}

\subsection{HV Simulator Module}

At each discrete control cycle with sampling time $\Delta t$, the HV simulator
reads human driver inputs and converts them into control commands for a
continuous-time kinematic single-track vehicle model.
Two categories of input devices are supported:
(i) continuous hardware devices, including a steering wheel and throttle/brake
pedals, and (ii) discrete keyboard inputs.
All input modalities are mapped to a unified set of normalized control signals,
which are subsequently processed by a common vehicle control abstraction.

\subsubsection{Input Signal Processing}
\paragraph{Device Normalization}
The framework supports two input devices---a steering-wheel-and-pedals rig (Logitech G923) and a keyboard---both of which are mapped to the same triplet $(p_{\mathrm{th}}, p_{\mathrm{br}}, s)$ used downstream by the vehicle-dynamics control. The hardware rig provides continuous measurements that are normalized linearly, whereas the keyboard provides only binary on/off signals that must be smoothed into continuous values by rate-limited integration.

\emph{Hardware.} The steering wheel angle and pedal travel lengths are measured directly. Let $\ell_{\mathrm{th}} \in [0,\ell_{\mathrm{th,max}}]$ and $\ell_{\mathrm{br}} \in [0,\ell_{\mathrm{br,max}}]$ denote the physical throttle and brake pedal travel lengths, and let $\theta_{\mathrm{sw}} \in [-\theta_{\mathrm{sw,max}},\theta_{\mathrm{sw,max}}]$ denote the steering wheel angle relative to its mechanical limits. These raw measurements are linearly mapped to
\begin{equation}
\begin{aligned}
p_{\mathrm{th}} &= \frac{\ell_{\mathrm{th}}}{\ell_{\mathrm{th,max}}} \in [0,1], \\
p_{\mathrm{br}} &= \frac{\ell_{\mathrm{br}}}{\ell_{\mathrm{br,max}}} \in [0,1], \\
s &= \frac{\theta_{\mathrm{sw}}}{\theta_{\mathrm{sw,max}}} \in [-1,1],
\end{aligned}
\end{equation}
without further conditioning.

\emph{Keyboard.} At each control cycle of period $\Delta t$, let $b_{\mathrm{th}}, b_{\mathrm{br}}, b_L, b_R \in \{0,1\}$ denote the press states of the throttle (\texttt{W}/\texttt{Up}), brake (\texttt{S}/\texttt{Down}/\texttt{Space}), left-steer (\texttt{A}/\texttt{Left}), and right-steer (\texttt{D}/\texttt{Right}) keys. Whenever both pedal keys are held simultaneously, braking takes precedence; we therefore define the effective throttle indicator $\beta_{\mathrm{th}} \triangleq b_{\mathrm{th}}(1-b_{\mathrm{br}})$, which equals $1$ only when the throttle is held and the brake is not. Each pedal signal then ramps toward $1$ at a key-specific rate while its (effective) key is active, and decays toward $0$ at a common release rate $r^-_p$ otherwise:
\begin{equation}
\begin{aligned}
p_{\mathrm{th}} &\leftarrow \operatorname{clip}_{[0,1]} \bigl( p_{\mathrm{th}} + \bigl[ r^{+}_{\mathrm{th}}\, \beta_{\mathrm{th}} - r^{-}_{p}\, (1 - \beta_{\mathrm{th}}) \bigr]\, \Delta t \bigr),\\
p_{\mathrm{br}} &\leftarrow \operatorname{clip}_{[0,1]} \bigl( p_{\mathrm{br}} + \bigl[ r^{+}_{\mathrm{br}}\, b_{\mathrm{br}} - r^{-}_{p}\, (1 - b_{\mathrm{br}}) \bigr]\, \Delta t \bigr).
\end{aligned}
\end{equation}
The steering signal accumulates toward $\pm s_{\max}$ at rate $r^{+}_{s}$ while a steering key is held, and decays back to zero at rate $r^{-}_{s}$ when neither is held:
\begin{equation}
s \leftarrow
\begin{cases}
\operatorname{clip}_{[-s_{\max},\, s_{\max}]} \bigl( s + r^{+}_{s}\, (b_L - b_R)\, \Delta t \bigr), & b_L + b_R \ge 1,\\[2pt]
\operatorname{sgn}(s)\, \max \bigl( 0,\, |s| - r^{-}_{s}\, \Delta t \bigr), & b_L = b_R = 0.
\end{cases}
\end{equation}
We use $r^+_{\mathrm{th}} = 1.8\,\mathrm{s^{-1}}$, $r^+_{\mathrm{br}} = 2.5\,\mathrm{s^{-1}}$, $r^-_p = 2.2\,\mathrm{s^{-1}}$, $r^+_s = 2.0\,\mathrm{s^{-1}}$, $r^-_s = 3.5\,\mathrm{s^{-1}}$, and $s_{\max} = 0.65$; the steering cap is set below the full normalized range to keep the emulated wheel within a comfortable manual span. Small deadbands ($10^{-3}$ for the pedals and $10^{-4}$ for the steering) are applied to suppress numerical noise.

The resulting triplet $(p_{\mathrm{th}}, p_{\mathrm{br}}, s)$ is consumed identically by the subsequent vehicle-dynamics control, regardless of which device produced it.

\subsubsection{Vehicle Dynamics Control}
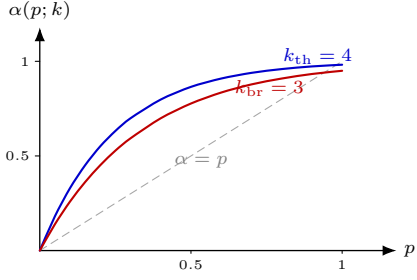
\begin{figure}[t]
\centering
\footnotesize
\begin{tikzpicture}[>=Latex, line cap=round, line join=round, x=4cm, y=2.5cm]
  \draw[->, line width=0.5pt] (0,0) -- (1.18,0) node[right] {\scriptsize $p$};
  \draw[->, line width=0.5pt] (0,0) -- (0,1.18) node[above] {\scriptsize $\alpha(p;k)$};

  \draw (0.5,0) -- (0.5,-0.015); \node[below=1pt, font=\tiny] at (0.5,0) {$0.5$};
  \draw (1.0,0) -- (1.0,-0.015); \node[below=1pt, font=\tiny] at (1.0,0) {$1$};
  \draw (0,0.5) -- (-0.015,0.5); \node[left=1pt, font=\tiny] at (0,0.5) {$0.5$};
  \draw (0,1.0) -- (-0.015,1.0); \node[left=1pt, font=\tiny] at (0,1.0) {$1$};

  \draw[densely dashed, gray!70] (0,0) -- (1,1);
  \node[gray, anchor=south west, font=\scriptsize] at (0.42,0.40) {$\alpha = p$};

  \draw[blue!80!black, thick, smooth] plot coordinates {
    (0,0) (0.05,0.181) (0.10,0.330) (0.15,0.451) (0.20,0.551)
    (0.25,0.632) (0.30,0.699) (0.40,0.798) (0.50,0.865)
    (0.60,0.909) (0.70,0.939) (0.80,0.959) (0.90,0.973) (1.0,0.982)
  };
  \node[blue!80!black, anchor=west, font=\scriptsize] at (0.78,1.03) {$k_{\mathrm{th}} = 4$};

  \draw[red!75!black, thick, smooth] plot coordinates {
    (0,0) (0.05,0.139) (0.10,0.259) (0.15,0.362) (0.20,0.451)
    (0.25,0.528) (0.30,0.593) (0.40,0.699) (0.50,0.777)
    (0.60,0.835) (0.70,0.878) (0.80,0.909) (0.90,0.933) (1.0,0.950)
  };
  \node[red!75!black, anchor=west, font=\scriptsize] at (0.62,0.86) {$k_{\mathrm{br}} = 3$};
\end{tikzpicture}
\caption{Exponential pedal shaping $\alpha(p;k) = 1 - \mathrm{e}^{-kp}$ used by the HV longitudinal controller, with the throttle gain $k_{\mathrm{th}} = 4$ (blue) and the brake gain $k_{\mathrm{br}} = 3$ (red); the linear identity $\alpha = p$ (dashed) is shown for comparison. A larger shape parameter $k$ produces a steeper initial slope and earlier saturation toward $\alpha = 1$, which gives finer-grained control near rest while still approaching full drive/brake authority at larger pedal travel.}
\label{fig:pedal_curves}
\end{figure}

\paragraph{Longitudinal Control}
The normalized throttle and brake pedal positions $p_{\mathrm{th}}, p_{\mathrm{br}} \in [0,1]$ produced by Device Normalization are each mapped to a dimensionless \emph{shaped pedal response} $\alpha \in [0,1)$ through a smooth saturating function
\begin{equation}
\begin{aligned}
\alpha(p; k) &= 1 - \mathrm{e}^{-kp}, \\
\alpha_{\mathrm{th}} &= \alpha(p_{\mathrm{th}};\, k_{\mathrm{th}}), \\
\alpha_{\mathrm{br}} &= \alpha(p_{\mathrm{br}};\, k_{\mathrm{br}}),
\end{aligned}
\end{equation}
where the \emph{shape parameter} $k > 0$ controls the steepness of the curve near the origin: a larger $k$ produces a greater response at small pedal travel and saturates toward $\alpha = 1$ more quickly. Because $p \in [0,1]$ and $k > 0$ imply $kp \geq 0$, the factor $\mathrm{e}^{-kp}$ lies in $(0,1]$, so the response is bounded as $\alpha \in [0,1)$ by construction---with $\alpha = 0$ at zero pedal travel and $\alpha = 1 - \mathrm{e}^{-k}$ at full travel---and no explicit clamping is required. We use $k_{\mathrm{th}} = 4$ for the throttle and $k_{\mathrm{br}} = 3$ for the brake; the two curves are shown together with the linear identity $\alpha = p$ in Fig.~\ref{fig:pedal_curves}.

The exponential form is adopted because (i)~it is monotone, smooth, and differentiable on $[0,1]$, which is convenient for downstream filtering and saturation; (ii)~it yields finer-grained control at small pedal travel, the region in which drivers concentrate most of their pedal modulation~\cite{treiber2000congested}, than the linear mapping $\alpha = p$; and (iii)~it abstracts away detailed powertrain and brake-actuator transfer dynamics~\cite{rajamani2012vehicle}, which are not the focus of this work.

Given the current vehicle speed $v$, the maximum tractive force is constrained by
both engine power and tire--road friction~\cite{rajamani2012vehicle},
\begin{equation}
F_{\max}(v)=\min\!\left(\frac{P_{\max}}{|v|+\varepsilon_v},\;\mu m g\right),
\end{equation}
where $P_{\max}$ denotes the peak engine power, $\mu$ the tire--road friction
coefficient, $m$ the vehicle mass, $g$ the gravitational acceleration, and
$\varepsilon_v>0$ a small regularization constant preventing singularity at low
speeds.

Resistive forces consist of aerodynamic drag and rolling resistance
\cite{gillespie1992fundamentals},
\begin{equation}
\begin{aligned}
F_{\mathrm{drag}} &= \tfrac{1}{2}\, \rho\, C_D\, A_f\, |v|\, v, \\
F_{\mathrm{roll}} &= m g\, C_{rr}\, \operatorname{sgn}(v),
\end{aligned}
\end{equation}
where $\rho$ denotes air density, $C_D$ the drag coefficient, $A_f$ the frontal
area, and $C_{rr}$ the rolling resistance coefficient.

The resulting drive and brake forces are given by
\begin{equation}
\begin{aligned}
F_{\mathrm{drive}} &= \alpha_{\mathrm{th}}\, F_{\max}(v), \\
F_{\mathrm{brake}} &= \alpha_{\mathrm{br}}\, F_{\mathrm{brake,max}},
\end{aligned}
\end{equation}
where $F_{\mathrm{brake,max}}$ denotes the maximum achievable braking force.
The commanded longitudinal acceleration is then computed as
\begin{equation}
a_{\mathrm{cmd}}=\mathrm{sat}_{[-a_{\max},a_{\max}]}\!\left(
\frac{F_{\mathrm{drive}}-F_{\mathrm{brake}}-F_{\mathrm{drag}}-F_{\mathrm{roll}}}{m}
\right),
\end{equation}
where $a_{\max}>0$ denotes the maximum admissible longitudinal acceleration and
$\mathrm{sat}_{[a,b]}(\cdot)$ clamps its argument to the interval $[a,b]$.

To emulate drivetrain and brake dynamics, the commanded acceleration is filtered
by a first-order lag with time constant $\tau_{\mathrm{long}} > 0$,
\begin{equation}
\dot a = \frac{1}{\tau_{\mathrm{long}}}\bigl(a_{\mathrm{cmd}} - a\bigr)
\;\;\Longleftrightarrow\;\;
a_{n+1} = a_n + \frac{\Delta t}{\tau_{\mathrm{long}}}\bigl(a_{\mathrm{cmd},n} - a_n\bigr),
\end{equation}
where $n$ denotes the discrete simulation step index and $a$ is the filtered (delivered) longitudinal acceleration.

\paragraph{Lateral Control}

The normalized steering input $s\in[-1,1]$ is first mapped to a steering-wheel
angle $\theta_{\mathrm{sw}}=s\,\theta_{\mathrm{sw,max}}$, where $\theta_{\mathrm{sw,max}}=7.85\,\mathrm{rad}$ ($450^\circ$).
The corresponding \emph{target front-wheel steering angle} $\delta^\star$ is
obtained by applying a variable steering ratio and enforcing physical limits.
The target steering angle is computed as
\begin{equation}
\delta^\star=\mathrm{sat}_{[-\delta_{\max},\delta_{\max}]}\!\left(
\frac{\theta_{\mathrm{sw}}}{R(v,\theta_{\mathrm{sw}})}
\right),
\end{equation}
where $R(v, \theta_{\mathrm{sw}})$ is a calibrated steering ratio obtained via lookup tables indexed by speed and steering angle, and $\delta_{\max} = 0.91\,\mathrm{rad}$ represents the physical front-wheel steering angle limit.

The steering-rate command supplied to the vehicle model is approximated by the
discrete-time derivative of the steering angle,
\begin{equation}
\dot\delta_{\mathrm{cmd}}=\frac{\delta^\star-\delta}{\Delta t},
\end{equation}
where $\delta$ denotes the current front-wheel steering angle and $\Delta t$ the
simulation sampling time.
This formulation is consistent with control interfaces commonly adopted in
motion-planning benchmarks such as CommonRoad~\cite{althoff2017commonroad}.

\paragraph{Trajectory Generation}
The HV dynamics are driven by the steering-rate command $\dot\delta_{\mathrm{cmd}}$ and the filtered longitudinal acceleration $a$, and are numerically integrated at each control cycle over the interval $[t, t+\Delta t]$.
Repeating this procedure yields an HV trajectory consistent with the driver's inputs, while respecting physical limits (power and friction constraints, braking capacity, and steering bounds) and ensuring smoothness through input ramping and acceleration lag.

\subsection{Drivability Detection Module}
To monitor safety, the simulator optionally performs collision and road-compliance
checks for the HV using the CommonRoad drivability-checker
package~\cite{pek2020commonroad}.
Checks are evaluated at each control cycle using the current vehicle states
only.

Let the HV simulator state at scenario time index $k$ be
$x_{h}^{\mathrm{sim}} = [x_{h}, y_{h}, \delta_{h}, v_{h}, \psi_{h}]^\top$,
consistent with the notation of the Low-Level Simulation Module.
This state is mapped to a CommonRoad kinematic single-track (KS) state
$x_{h}^{k} = (p_{h}^{k}, v_{h}^{k}, \psi_{h}^{k}, \delta_{h}^{k})$
by applying the same sim-to-scenario rigid transform $T_{s\rightarrow c}$,
\begin{equation}
\begin{aligned}
p_{h}^{k} &= R\, p_{h}^{\mathrm{sim}} + \mathbf{d}, \\
\psi_{h}^{k} &= \mathrm{wrap}_\pi\!\left( \psi_{h}^{\mathrm{sim}} + \Delta\psi \right),
\end{aligned}
\end{equation}
where $R$, $\mathbf{d}$, and $\Delta\psi$ are the rotation, translation, and yaw offset of $T_{s\rightarrow c}$ defined by the Frame-Alignment Module.
The index $k$ corresponds to the nearest CommonRoad time step consistent with
the simulator time and scenario sampling period.

\paragraph{Collision Detection}
Each vehicle is approximated by a rectangle of length $l$ and width $w$.
The HV footprint is tested against a collision checker that aggregates
static obstacles from the scenario and dynamic obstacles representing all ego
vehicles at the same time step.
A collision is declared if the checker reports an intersection between the
HV and any obstacle.

\paragraph{Road-Compliance Detection}
Road boundaries are converted into a collision object derived from the lanelet
network.
The HV is considered off-road if its rectangular footprint collides
with this boundary object.

The drivability status is declared valid if neither collision nor off-road
violation is detected; these checks can be enabled or disabled independently via
configuration flags.
Violations are reported in the simulator and can trigger a stop request for the
HV to ensure safe termination.

\subsection{Visualisation Module}

\begin{figure}[t]
  \centering
  \includegraphics[width=\linewidth]{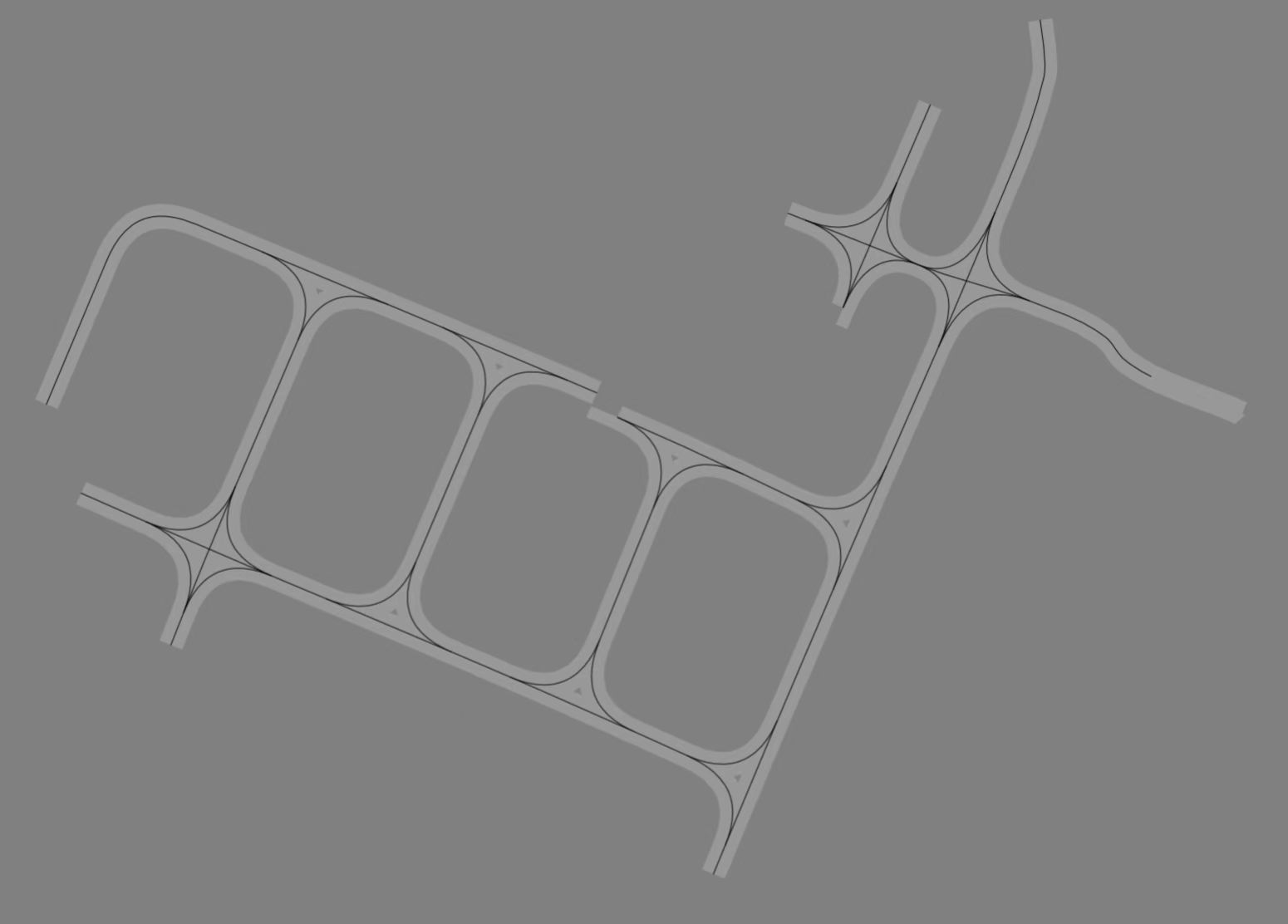}
  \caption{Example road-network rendering in the visualisation module.}
  \label{fig:road_network_example}
\end{figure}

The visualisation module renders the lanelet-based road network from the CommonRoad scenario, together with the current poses of the HV and the AV, into a live simulation window, providing the human driver with real-time visual feedback during the simulation.
The road network is given by left and right lane boundaries represented as
polylines in a global metric map frame.
For visual rendering purposes, these geometric primitives are converted into
filled lane surfaces and boundary curves within a fixed world window of size
$W\times H$ (in meters).
Fig.~\ref{fig:road_network_example} shows an example road-network rendering.

\subsection{Scenario Generation Module}
This module converts runtime simulation data into standardized
\textbf{CommonRoad scenario files}. It processes time-series state logs,
including position, velocity, and orientation of all agents, and reconstructs
the dynamic traffic environment in the CommonRoad XML format. By mapping the
recorded trajectories of both AVs and HVs onto the
underlying road network, the module generates high-fidelity replayable
CommonRoad scenarios. These standardized outputs support offline verification,
frame-by-frame visualization of interactions, and the sharing of representative
interaction cases without requiring the original simulation environment.

\setlength{\dblfloatsep}{1pt}
\setlength{\dbltextfloatsep}{1pt}

\section{Evaluation}
\label{sec:evaluation}
This section provides a qualitative evaluation of the proposed simulation framework
 by replaying a recorded HV and analyzing the resulting interaction with an AV controlled by different planners.
The simulator runs at a fixed time step of $\Delta t=0.01\,\mathrm{s}$, while 
figures report representative state snapshots at coarser intervals for readability.

\subsection{Single-AV Simulation with Different Planners}
We consider a single-AV scenario in which an HV is
actively controlled via keyboard input to exhibit aggressive interactive behaviors toward the AV.
For each planning policy under evaluation, the HV applies a consistent set of
challenging maneuvers, including sudden lane changes, intentional stopping in
front of the AV, and deliberate obstruction of the AV's intended path.

\subsubsection{The IDM Planner}

\begin{figure*}[!t]
    \centering
    \setlength{\abovecaptionskip}{3pt}
    \setlength{\belowcaptionskip}{-10pt}
    \subfloat[$t=0.00\,\mathrm{s}$]{\includegraphics[width=0.32\textwidth]{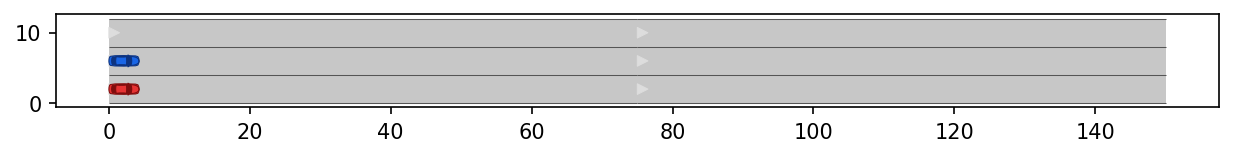}}\hfill\subfloat[$t=0.20\,\mathrm{s}$]{\includegraphics[width=0.32\textwidth]{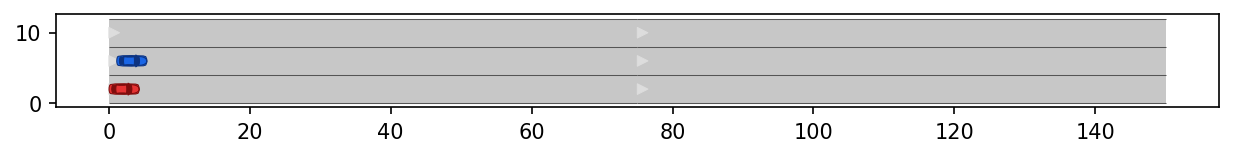}}\hfill\subfloat[$t=0.40\,\mathrm{s}$]{\includegraphics[width=0.32\textwidth]{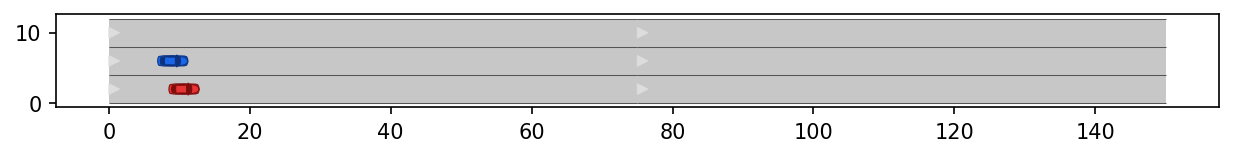}}\\[-2ex]
    \subfloat[$t=0.60\,\mathrm{s}$]{\includegraphics[width=0.32\textwidth]{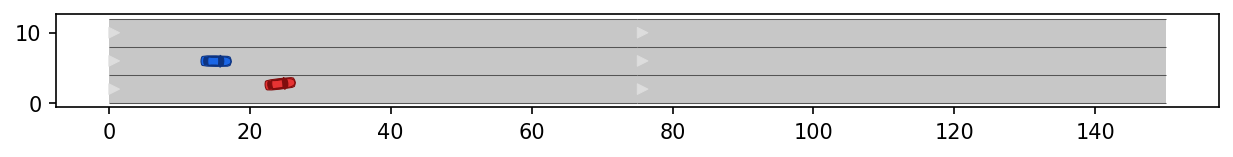}}\hfill\subfloat[$t=0.80\,\mathrm{s}$]{\includegraphics[width=0.32\textwidth]{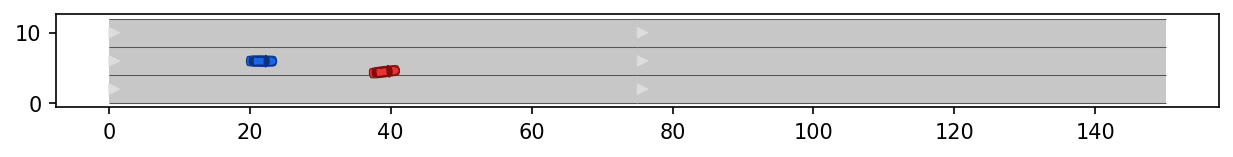}}\hfill\subfloat[$t=1.00\,\mathrm{s}$]{\includegraphics[width=0.32\textwidth]{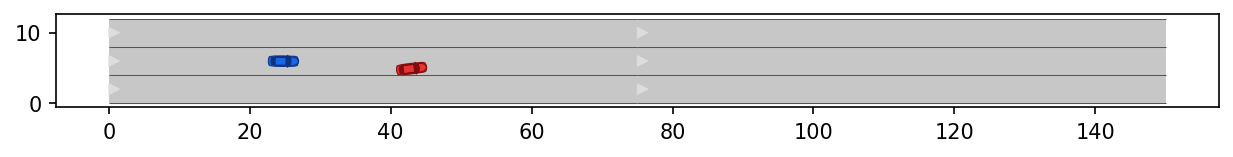}}\\[-2ex]
    \subfloat[$t=1.20\,\mathrm{s}$]{\includegraphics[width=0.32\textwidth]{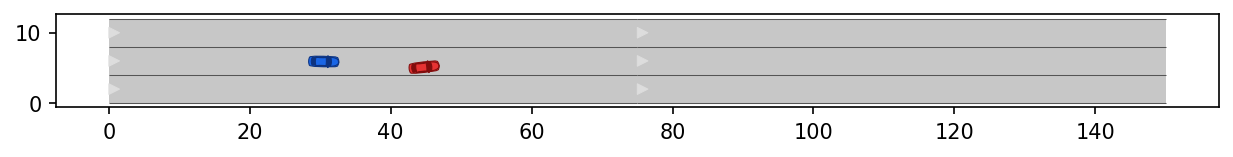}}\hfill\subfloat[$t=1.40\,\mathrm{s}$]{\includegraphics[width=0.32\textwidth]{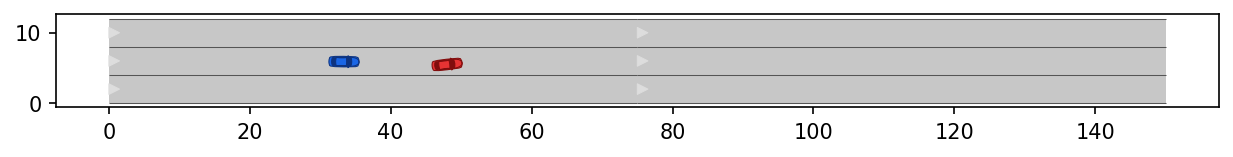}}\hfill\subfloat[$t=1.60\,\mathrm{s}$]{\includegraphics[width=0.32\textwidth]{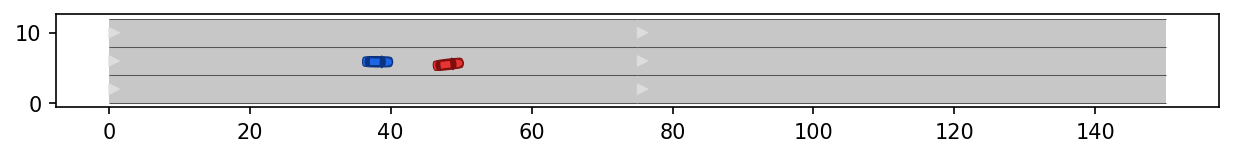}}\\[-2ex]
    \subfloat[$t=1.80\,\mathrm{s}$]{\includegraphics[width=0.32\textwidth]{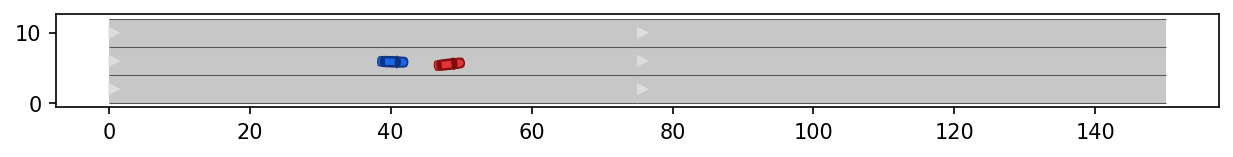}}\hfill\subfloat[$t=2.00\,\mathrm{s}$]{\includegraphics[width=0.32\textwidth]{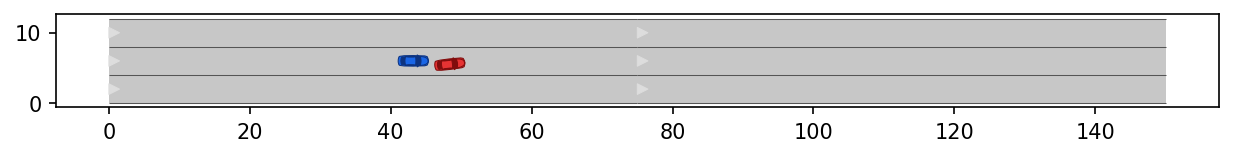}}\hfill\subfloat[$t=2.20\,\mathrm{s}$]{\includegraphics[width=0.32\textwidth]{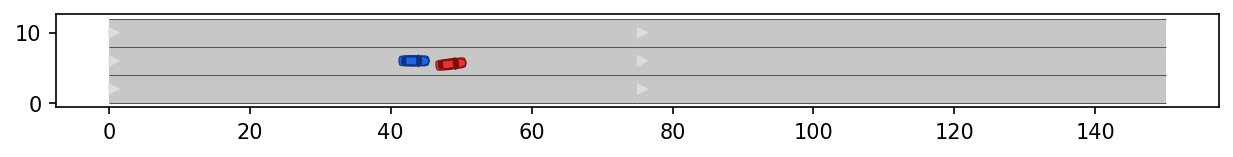}}\\[-2ex]
    \subfloat[$t=2.40\,\mathrm{s}$]{\includegraphics[width=0.32\textwidth]{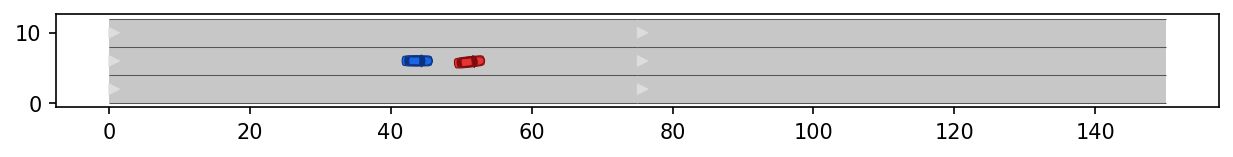}}\hfill\subfloat[$t=2.60\,\mathrm{s}$]{\includegraphics[width=0.32\textwidth]{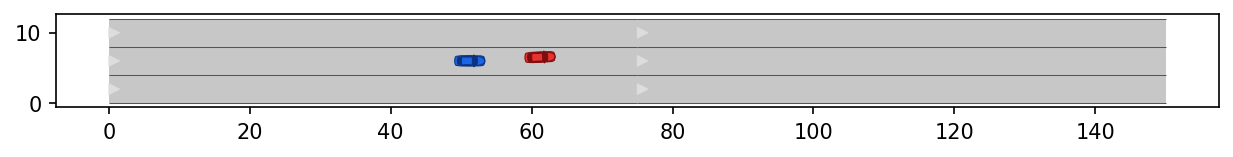}}\hfill\subfloat[$t=2.80\,\mathrm{s}$]{\includegraphics[width=0.32\textwidth]{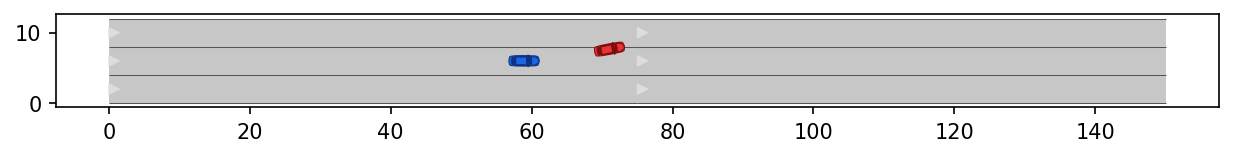}}\\[-2ex]
    \subfloat[$t=3.00\,\mathrm{s}$]{\includegraphics[width=0.32\textwidth]{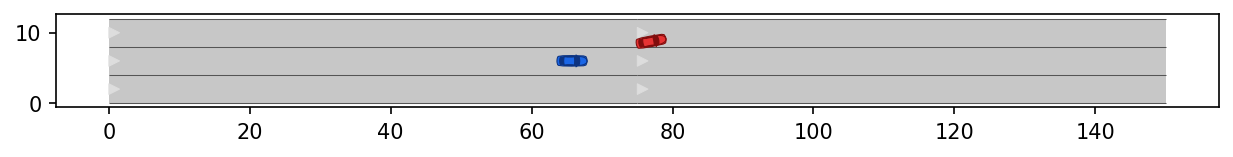}}\hfill\subfloat[$t=3.20\,\mathrm{s}$]{\includegraphics[width=0.32\textwidth]{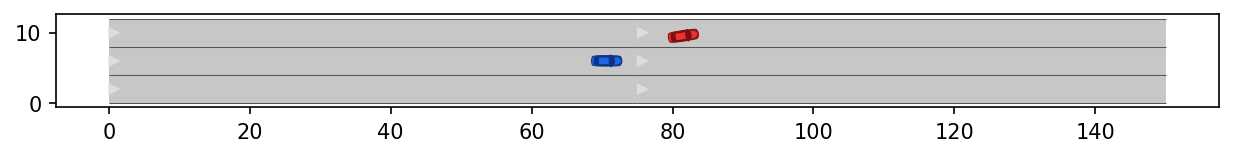}}\hfill\subfloat[$t=3.40\,\mathrm{s}$]{\includegraphics[width=0.32\textwidth]{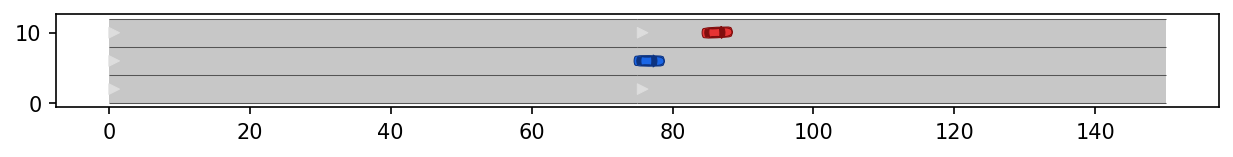}}\\
    \caption{IDM-planner per-frame state snapshots of the AV--HV interaction. The red car is the HV, and the blue car is the AV. Panels are shown at $0.2\,\mathrm{s}$ intervals (the simulator runs at $\Delta t=0.01\,\mathrm{s}$). The sequence shows the HV cutting in directly ahead of the AV and coming to a stop, after which the AV decelerates and stops to maintain a safe gap.}
    \label{fig:idm_states}
\end{figure*}

\begin{figure*}[!t]
    \centering
    \setlength{\abovecaptionskip}{3pt}
    \setlength{\belowcaptionskip}{-10pt}
    \subfloat[Heading angle]{\includegraphics[width=0.32\textwidth]{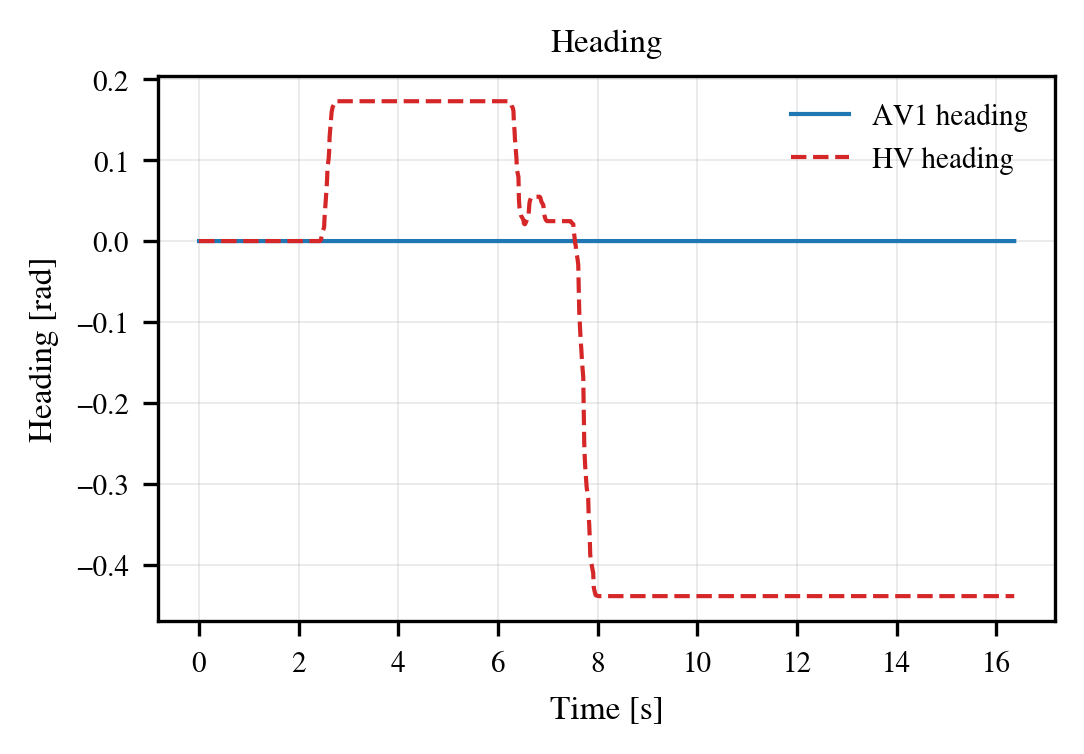}}
    \hfill
    \subfloat[Position]{\includegraphics[width=0.32\textwidth]{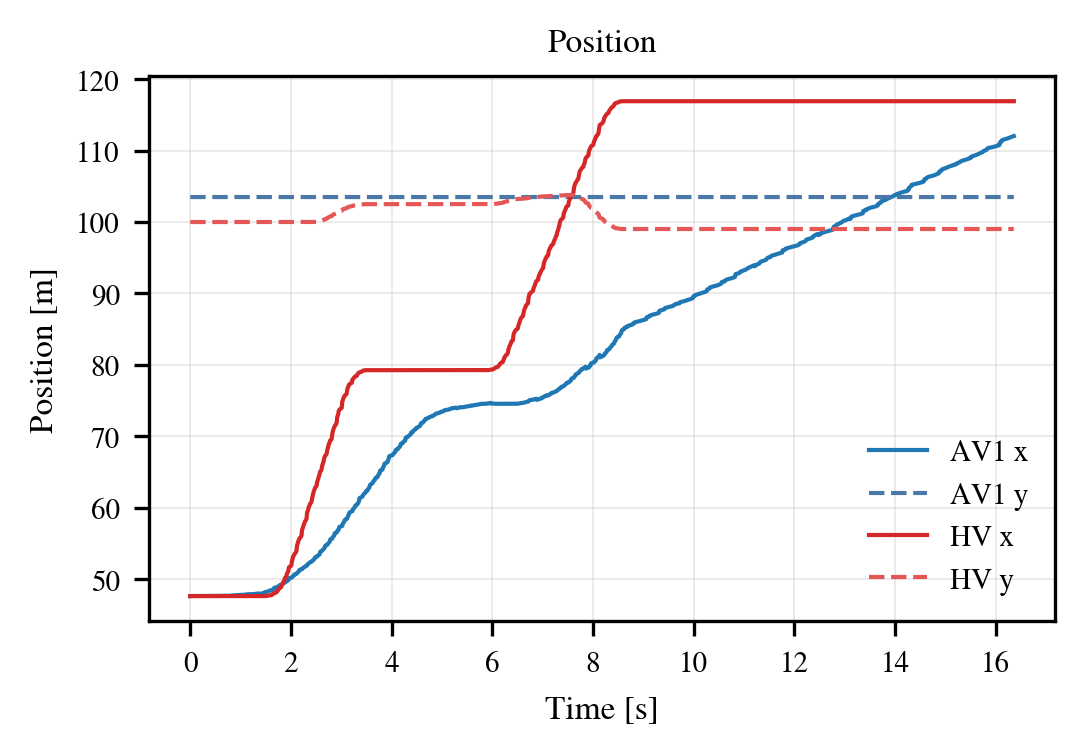}}
    \hfill
    \subfloat[Velocity]{\includegraphics[width=0.32\textwidth]{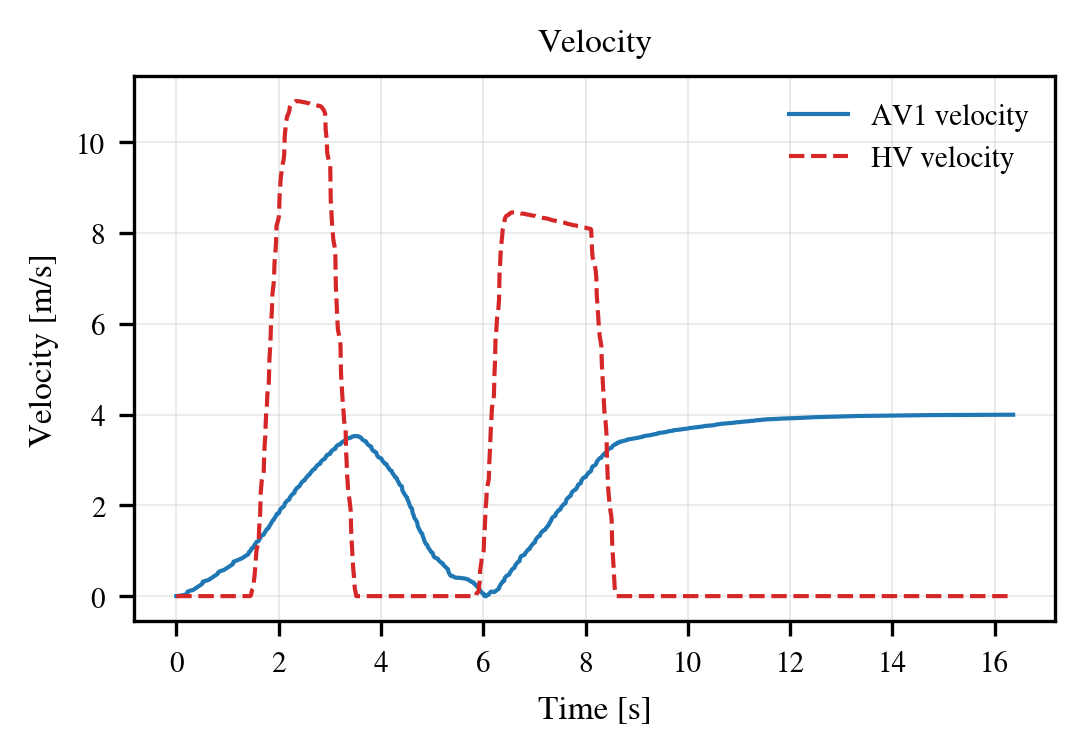}}
    \caption{Quantitative analysis of the IDM planner execution. (a) Heading angle profile, (b) Position trajectory, and (c) Velocity profile of the AV during the interaction.}
    \label{fig:idm_analysis}
\end{figure*}
Fig.~\ref{fig:idm_states} shows per-frame state snapshots for the IDM planner\textsuperscript{\ref{fn:idm_planner}}.
The sequence highlights a representative interaction in which the HV cuts in 
directly ahead of the AV and then stops; accordingly, the AV is driven to 
decelerate and come to a stop to maintain a safe gap.
Fig.~\ref{fig:idm_analysis} provides a quantitative analysis of this interaction, displaying the heading angle, position, and velocity profiles of the AV.

\subsubsection{The Reactive Planner}
\label{subsubsec:evaluation_reactive}

\begin{figure*}[!t]
    \centering
    \setlength{\abovecaptionskip}{3pt}
    \setlength{\belowcaptionskip}{-10pt}
    \subfloat[$t=0.00\,\mathrm{s}$]{\includegraphics[width=0.32\textwidth]{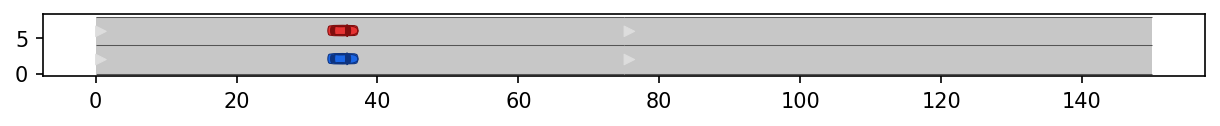}}\hfill\subfloat[$t=0.20\,\mathrm{s}$]{\includegraphics[width=0.32\textwidth]{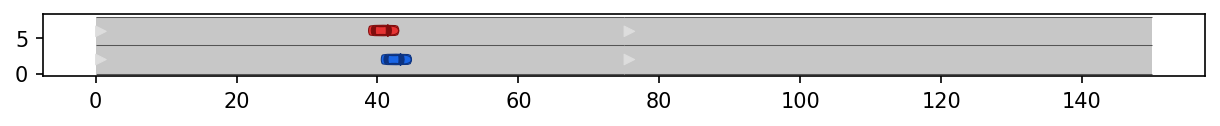}}\hfill\subfloat[$t=0.40\,\mathrm{s}$]{\includegraphics[width=0.32\textwidth]{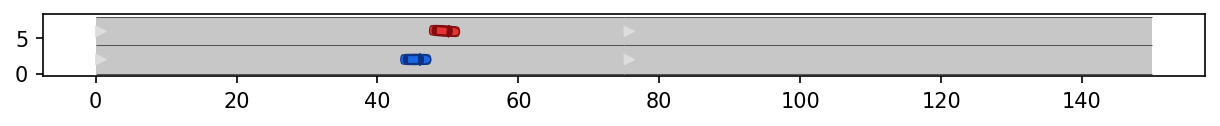}}\\[-2ex]
    \subfloat[$t=0.60\,\mathrm{s}$]{\includegraphics[width=0.32\textwidth]{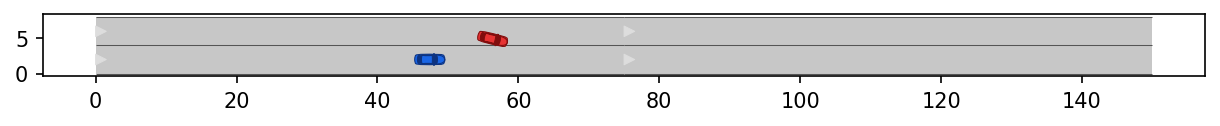}}\hfill\subfloat[$t=0.80\,\mathrm{s}$]{\includegraphics[width=0.32\textwidth]{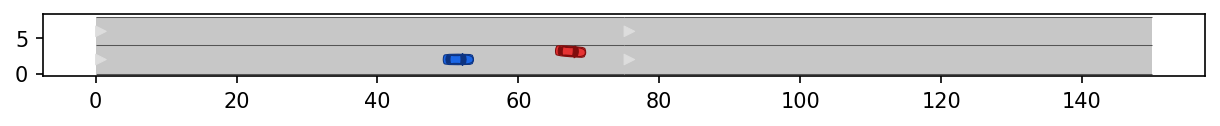}}\hfill\subfloat[$t=1.00\,\mathrm{s}$]{\includegraphics[width=0.32\textwidth]{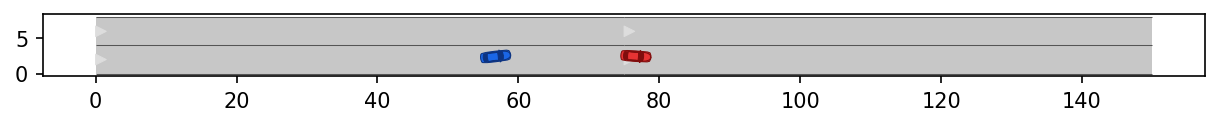}}\\[-2ex]
    \subfloat[$t=1.20\,\mathrm{s}$]{\includegraphics[width=0.32\textwidth]{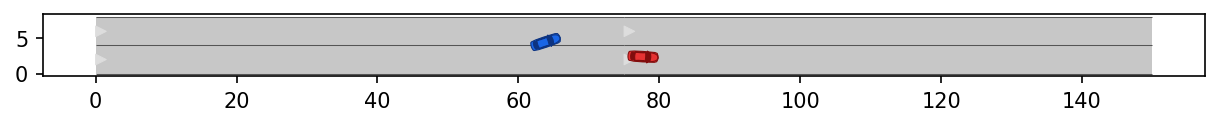}}\hfill\subfloat[$t=1.40\,\mathrm{s}$]{\includegraphics[width=0.32\textwidth]{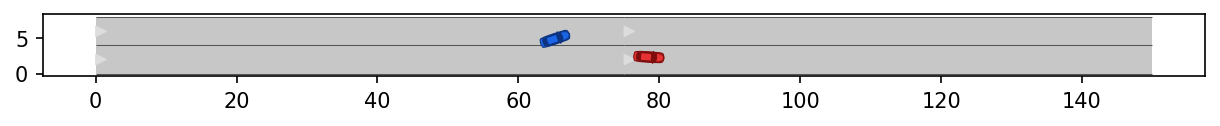}}\hfill\subfloat[$t=1.60\,\mathrm{s}$]{\includegraphics[width=0.32\textwidth]{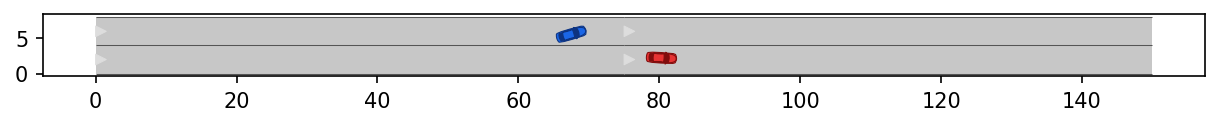}}\\[-2ex]
    \subfloat[$t=1.80\,\mathrm{s}$]{\includegraphics[width=0.32\textwidth]{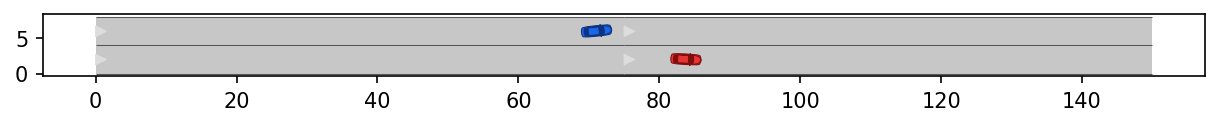}}\hfill\subfloat[$t=2.00\,\mathrm{s}$]{\includegraphics[width=0.32\textwidth]{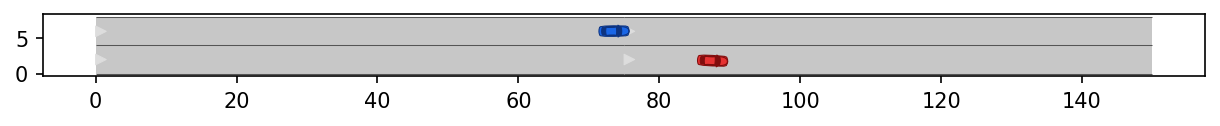}}\hfill\subfloat[$t=2.20\,\mathrm{s}$]{\includegraphics[width=0.32\textwidth]{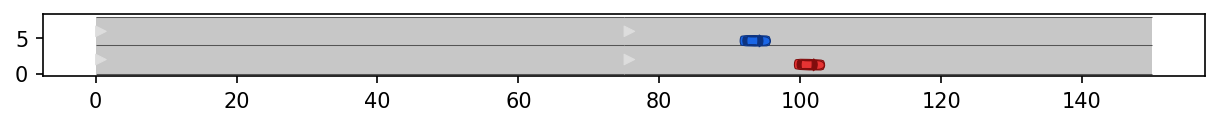}}\\[-2ex]
    \subfloat[$t=2.40\,\mathrm{s}$]{\includegraphics[width=0.32\textwidth]{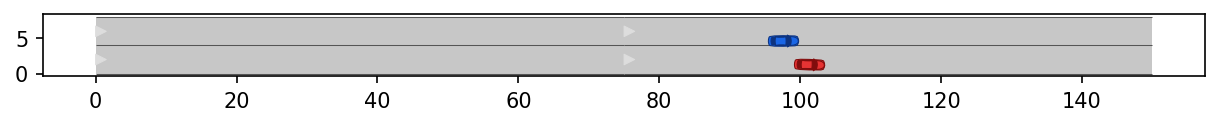}}\hfill\subfloat[$t=2.60\,\mathrm{s}$]{\includegraphics[width=0.32\textwidth]{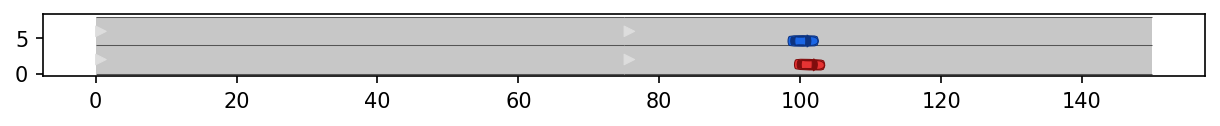}}\hfill\subfloat[$t=2.80\,\mathrm{s}$]{\includegraphics[width=0.32\textwidth]{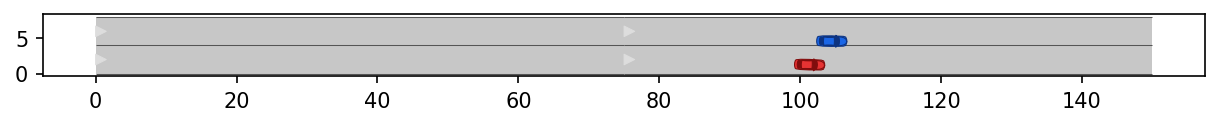}}\\[-2ex]
    \subfloat[$t=3.00\,\mathrm{s}$]{\includegraphics[width=0.32\textwidth]{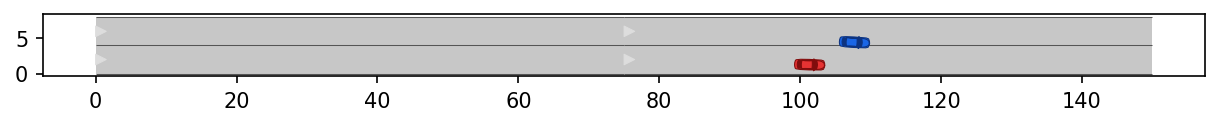}}\hfill\subfloat[$t=3.20\,\mathrm{s}$]{\includegraphics[width=0.32\textwidth]{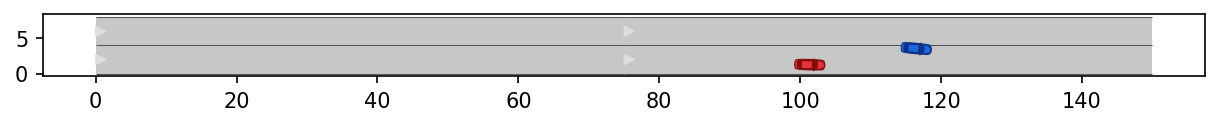}}\hfill\subfloat[$t=3.40\,\mathrm{s}$]{\includegraphics[width=0.32\textwidth]{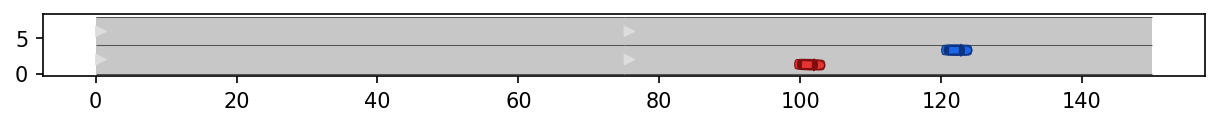}}\\
\caption{Reactive-planner simulation state snapshots of the AV--HV interaction. The red car is the HV, and the blue car is the AV. Panels are shown at $0.2\,\mathrm{s}$ intervals (the simulator runs at $\Delta t=0.01\,\mathrm{s}$). The sequence illustrates the AV executing a lane change while interacting with the HV.}
    \label{fig:reactive_states}
\end{figure*}

\begin{figure*}[!t]
    \centering
    \setlength{\abovecaptionskip}{3pt}
    \setlength{\belowcaptionskip}{-10pt}
    \subfloat[Heading angle]{\includegraphics[width=0.32\textwidth]{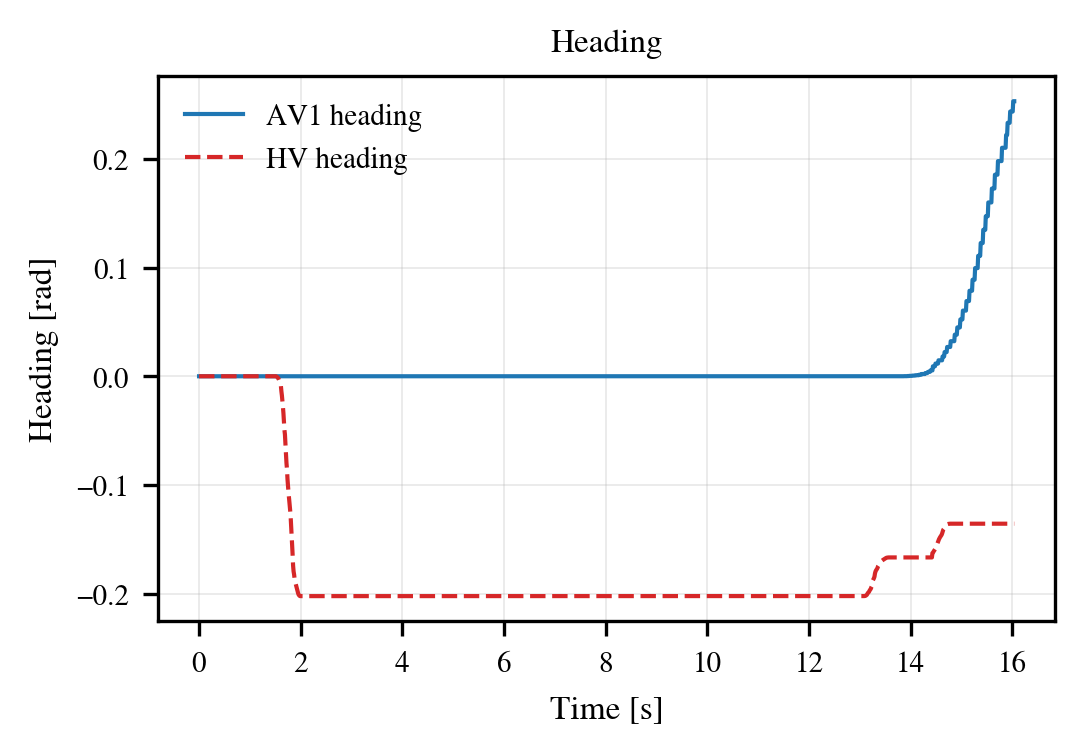}}
    \hfill
    \subfloat[Position]{\includegraphics[width=0.32\textwidth]{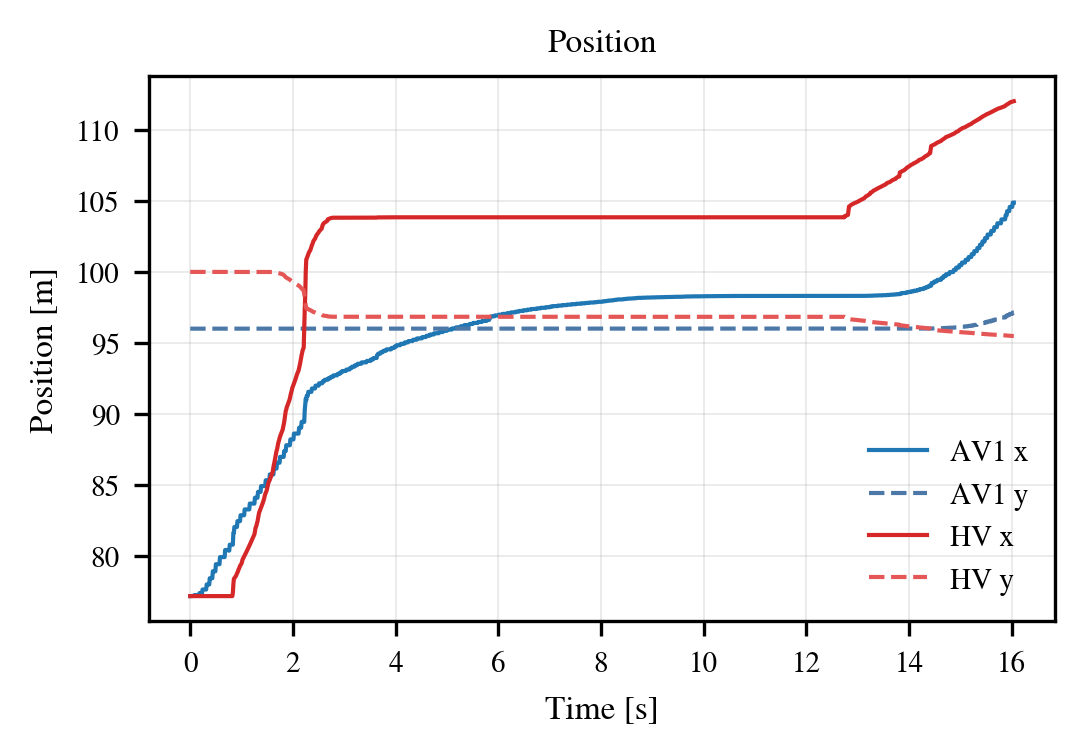}}
    \hfill
    \subfloat[Velocity]{\includegraphics[width=0.32\textwidth]{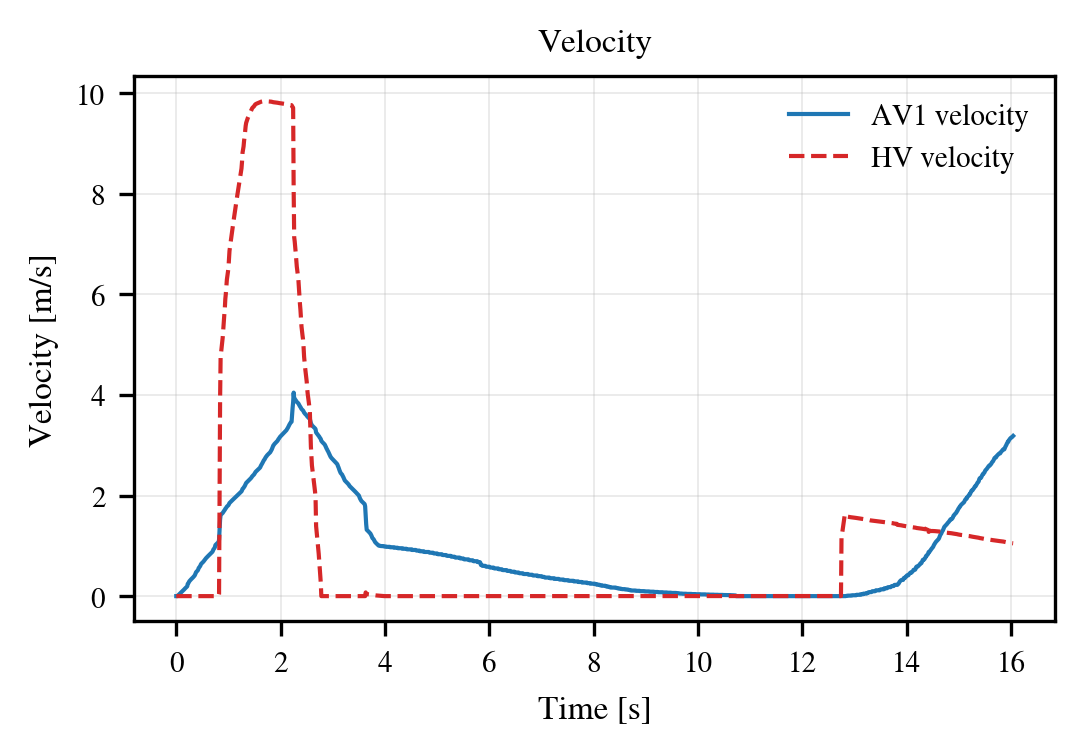}}
    \caption{Quantitative analysis of the Reactive planner execution. (a) Heading angle profile, (b) Position trajectory, and (c) Velocity profile of the AV during the interaction.}
    \label{fig:reactive_analysis}
\end{figure*}

We next replace the IDM planner with a sampling-based reactive planner.
Fig.~\ref{fig:reactive_states} shows the corresponding per-frame simulation states.
By leveraging a short-horizon prediction of the HV motion---assuming
constant velocity and heading over the prediction interval---the reactive planner
can proactively adjust its behavior, such as early deceleration or lane-change
initiation, in response to the anticipated HV trajectory.
As a result, the reactive planner executes a lane-change maneuver in this scenario,
leading to closer and more dynamic AV--HV interaction.
Fig.~\ref{fig:reactive_analysis} further illustrates the quantitative performance of the reactive planner, showing the evolution of its heading, position, and velocity during the maneuver.

\subsection{Multi-AV Simulation Capability}
\label{subsec:multi_av}

\begin{figure*}[!t]
    \centering
    \setlength{\abovecaptionskip}{3pt}
    \setlength{\belowcaptionskip}{-10pt}
    \subfloat[$t=0.00\,\mathrm{s}$]{\includegraphics[width=0.32\textwidth]{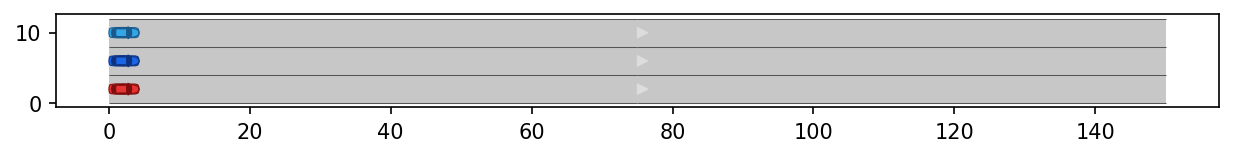}}\hfill\subfloat[$t=0.20\,\mathrm{s}$]{\includegraphics[width=0.32\textwidth]{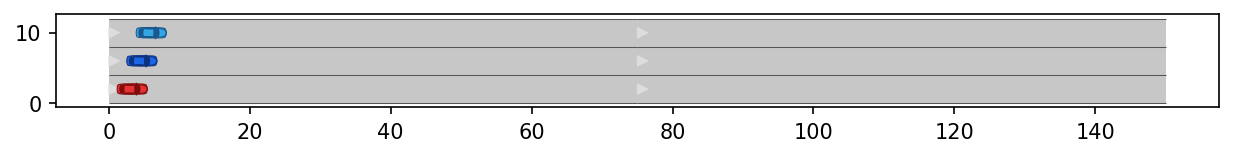}}\hfill\subfloat[$t=0.40\,\mathrm{s}$]{\includegraphics[width=0.32\textwidth]{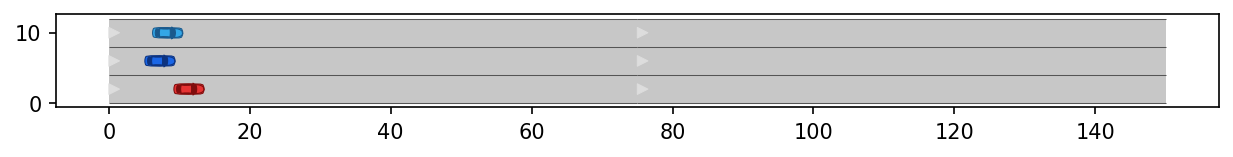}}\\[-2ex]
    \subfloat[$t=0.60\,\mathrm{s}$]{\includegraphics[width=0.32\textwidth]{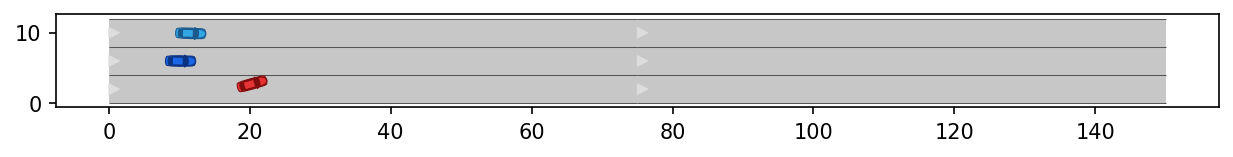}}\hfill\subfloat[$t=0.80\,\mathrm{s}$]{\includegraphics[width=0.32\textwidth]{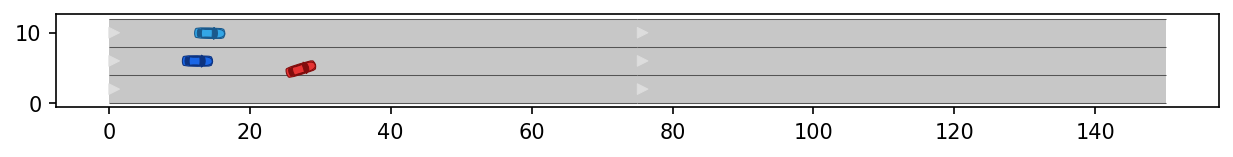}}\hfill\subfloat[$t=1.00\,\mathrm{s}$]{\includegraphics[width=0.32\textwidth]{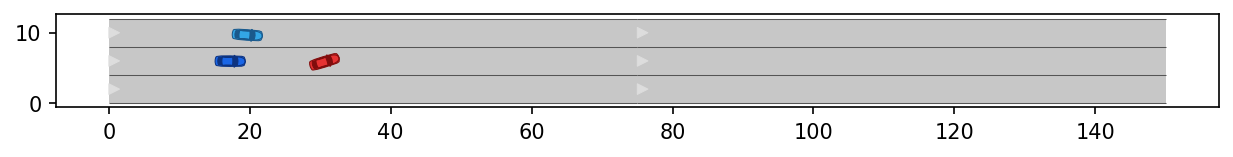}}\\[-2ex]
    \subfloat[$t=1.20\,\mathrm{s}$]{\includegraphics[width=0.32\textwidth]{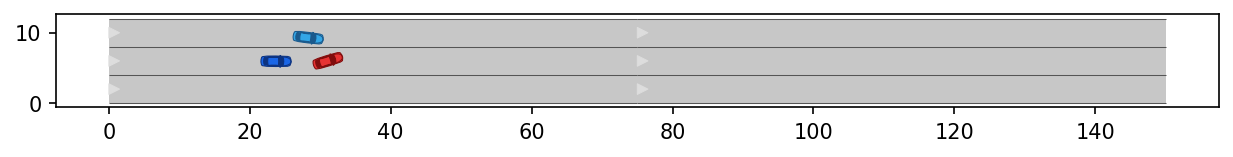}}\hfill\subfloat[$t=1.40\,\mathrm{s}$]{\includegraphics[width=0.32\textwidth]{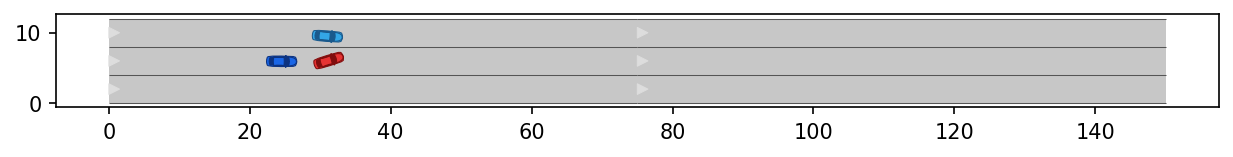}}\hfill\subfloat[$t=1.60\,\mathrm{s}$]{\includegraphics[width=0.32\textwidth]{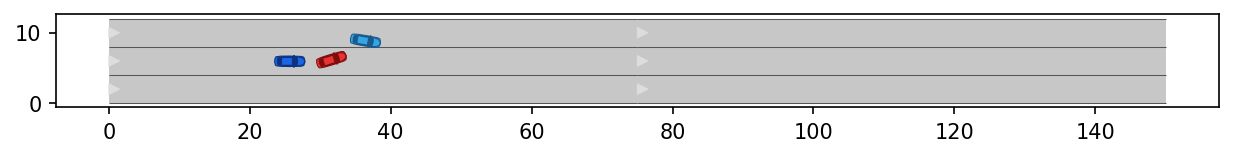}}\\[-2ex]
    \subfloat[$t=1.80\,\mathrm{s}$]{\includegraphics[width=0.32\textwidth]{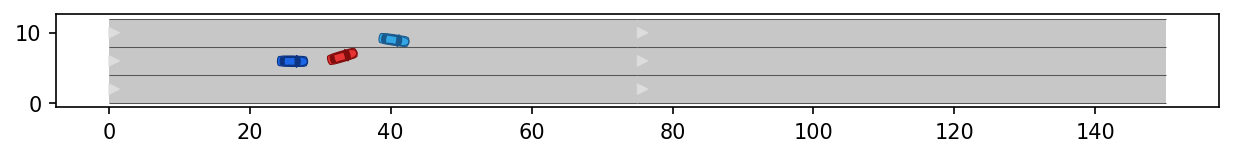}}\hfill\subfloat[$t=2.00\,\mathrm{s}$]{\includegraphics[width=0.32\textwidth]{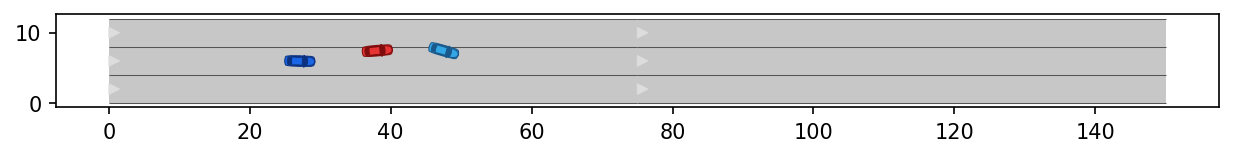}}\hfill\subfloat[$t=2.20\,\mathrm{s}$]{\includegraphics[width=0.32\textwidth]{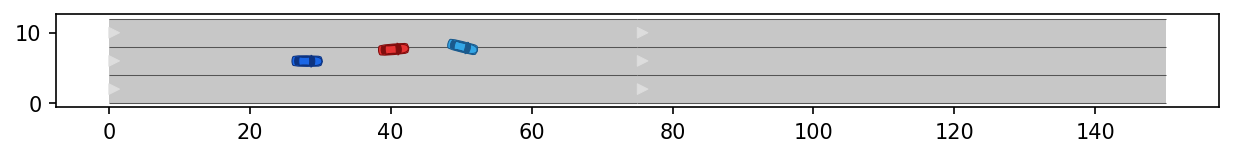}}\\[-2ex]
    \subfloat[$t=2.40\,\mathrm{s}$]{\includegraphics[width=0.32\textwidth]{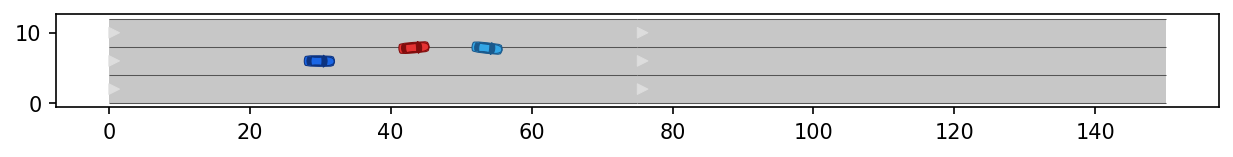}}\hfill\subfloat[$t=2.60\,\mathrm{s}$]{\includegraphics[width=0.32\textwidth]{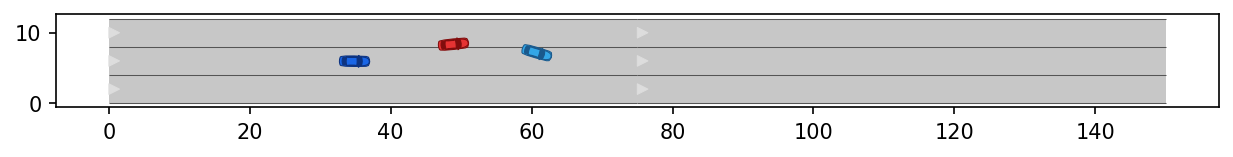}}\hfill\subfloat[$t=2.80\,\mathrm{s}$]{\includegraphics[width=0.32\textwidth]{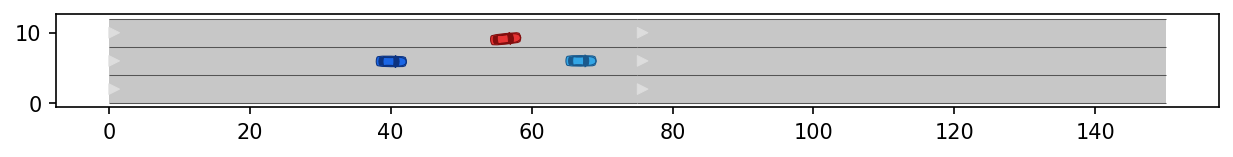}}\\[-2ex]
    \subfloat[$t=3.00\,\mathrm{s}$]{\includegraphics[width=0.32\textwidth]{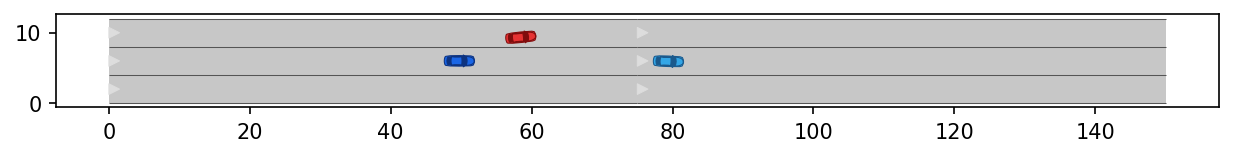}}\hfill\subfloat[$t=3.20\,\mathrm{s}$]{\includegraphics[width=0.32\textwidth]{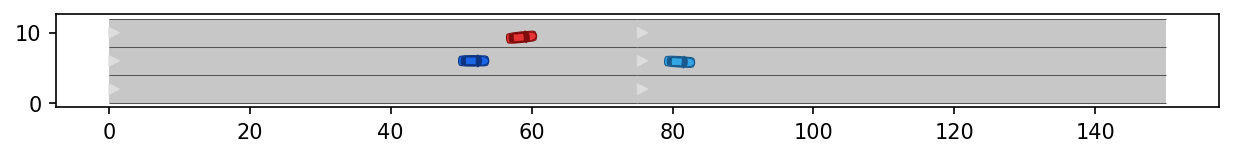}}\hfill\subfloat[$t=3.40\,\mathrm{s}$]{\includegraphics[width=0.32\textwidth]{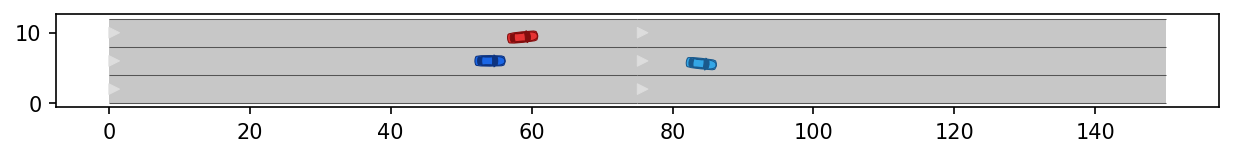}}\\
    \caption{Per-frame simulation states in a multi-AV scenario. The red car is the HV, and the blue cars are the AVs.
    Multiple AVs are simulated concurrently within the same road network. Panels are shown at $0.2\,\mathrm{s}$ intervals.}
    \label{fig:multi_av_states}
\end{figure*}

\begin{figure*}[!t]
    \centering
    \setlength{\abovecaptionskip}{3pt}
    \setlength{\belowcaptionskip}{-10pt}
    \subfloat[Heading angle]{\includegraphics[width=0.32\textwidth]{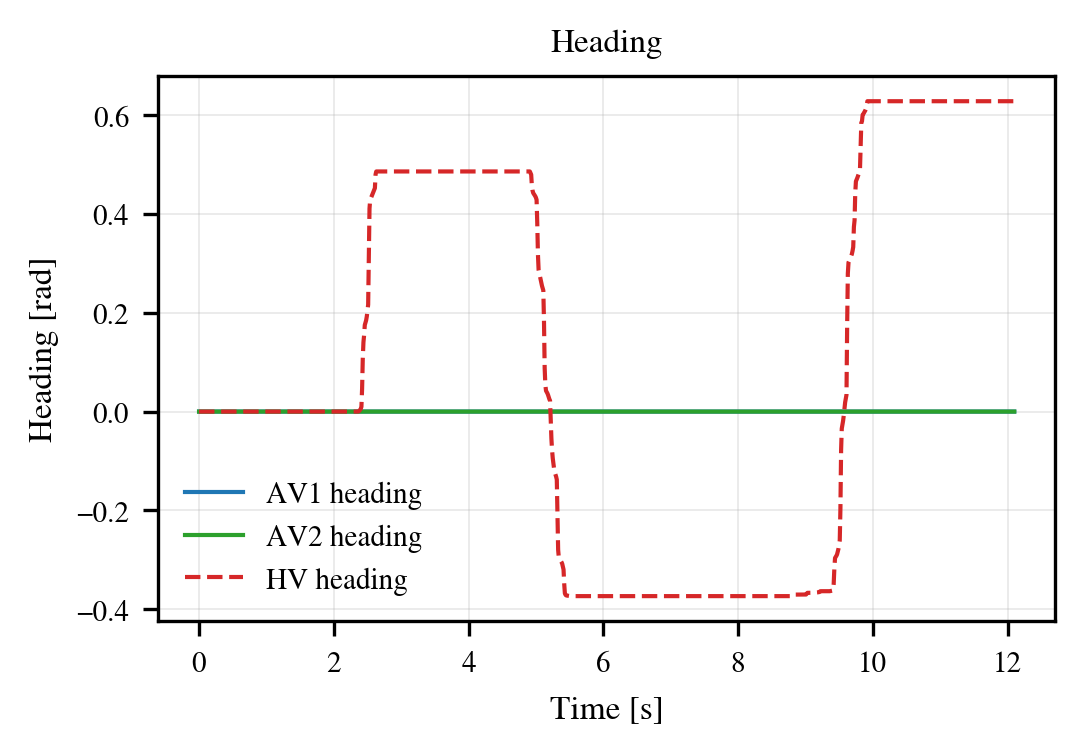}}
    \hfill
    \subfloat[Position]{\includegraphics[width=0.32\textwidth]{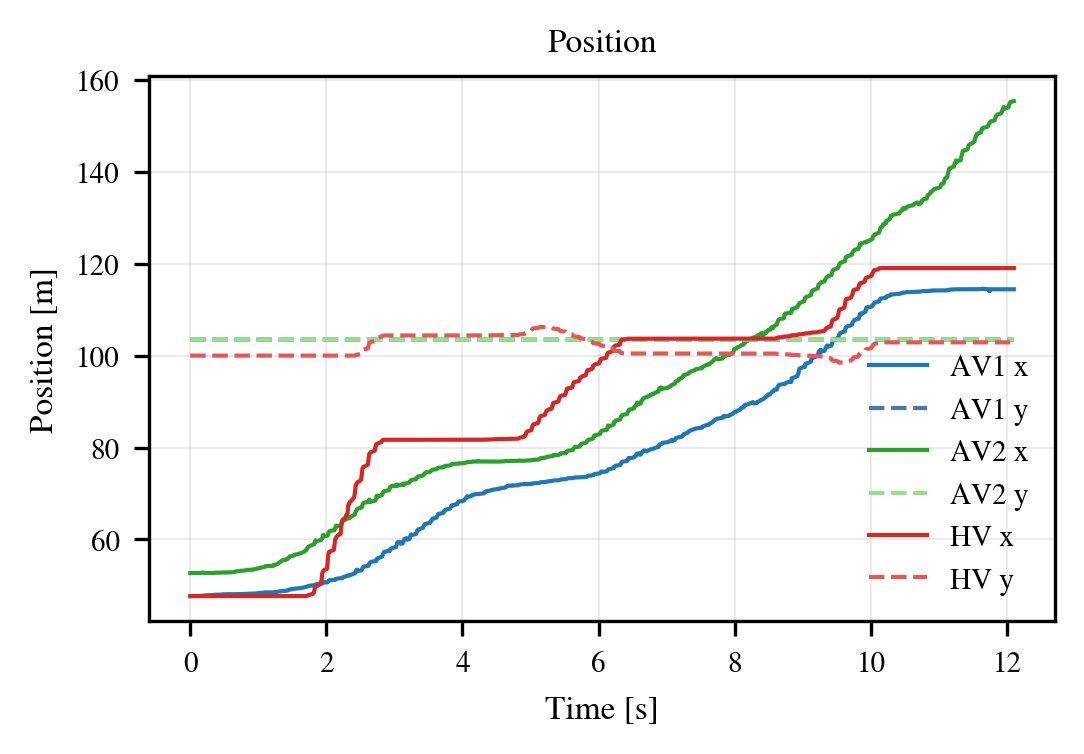}}
    \hfill
    \subfloat[Velocity]{\includegraphics[width=0.32\textwidth]{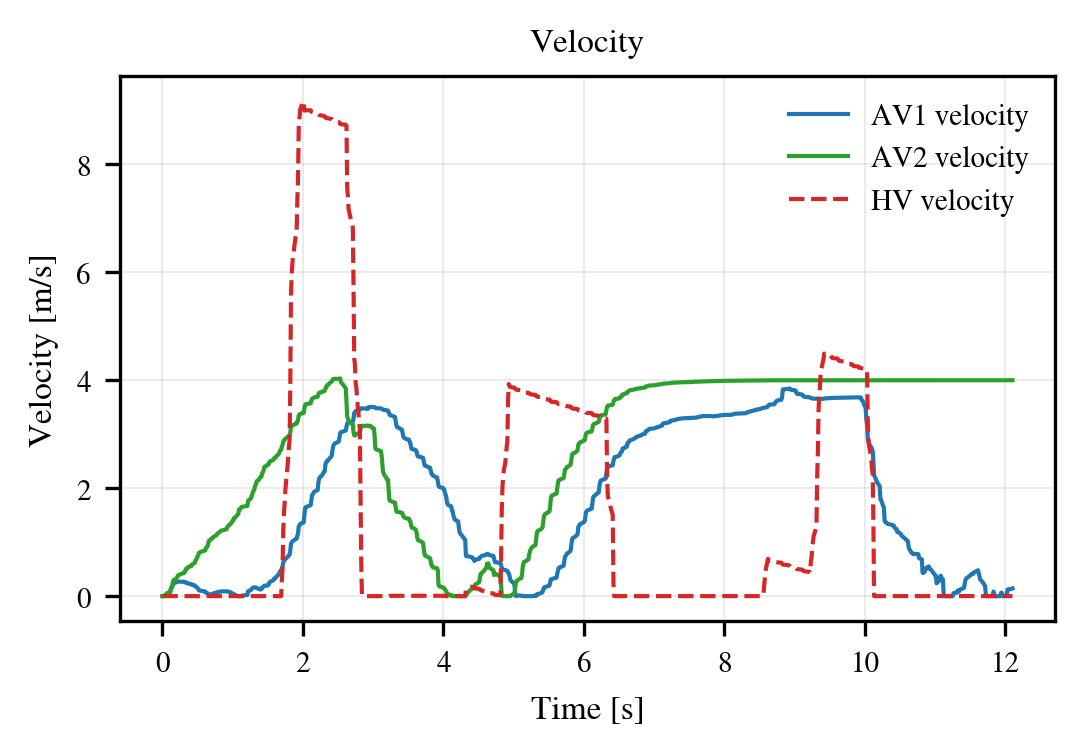}}
    \caption{Quantitative analysis of the Multi-AV scenario. (a) Heading angle profiles, (b) Position trajectories, and (c) Velocity profiles of the two AVs and the HV during the interaction.}
    \label{fig:multi_av_analysis}
\end{figure*}

To demonstrate the scalability of the simulation framework, we further evaluate a multi-AV setup, where two AVs are simulated concurrently within the same CommonRoad~\cite{althoff2017commonroad} scenario. Each AV is controlled by an independent planner instance, while sharing a common simulation clock and road network representation. Fig.~\ref{fig:multi_av_states} presents a sequence of per-frame simulation states for a representative multi-AV scenario. The results show that the framework can consistently update and visualise the states of multiple vehicles without temporal inconsistencies, confirming its suitability for multi-agent simulation and evaluation. Fig.~\ref{fig:multi_av_analysis} provides a quantitative analysis of the multi-AV scenario, displaying the heading angle, position, and velocity profiles for the two AVs and the HV. The consistent and smooth evolution of these states for all agents further validates the framework's capability to handle multi-agent dynamics and interactions effectively.

\subsection{Timing Synchronization Performance}
\label{subsec:timing_eval}

\begin{figure*}[!b]
    \centering
    \subfloat[Proposed: Timing Progression]{\includegraphics[width=0.48\linewidth]{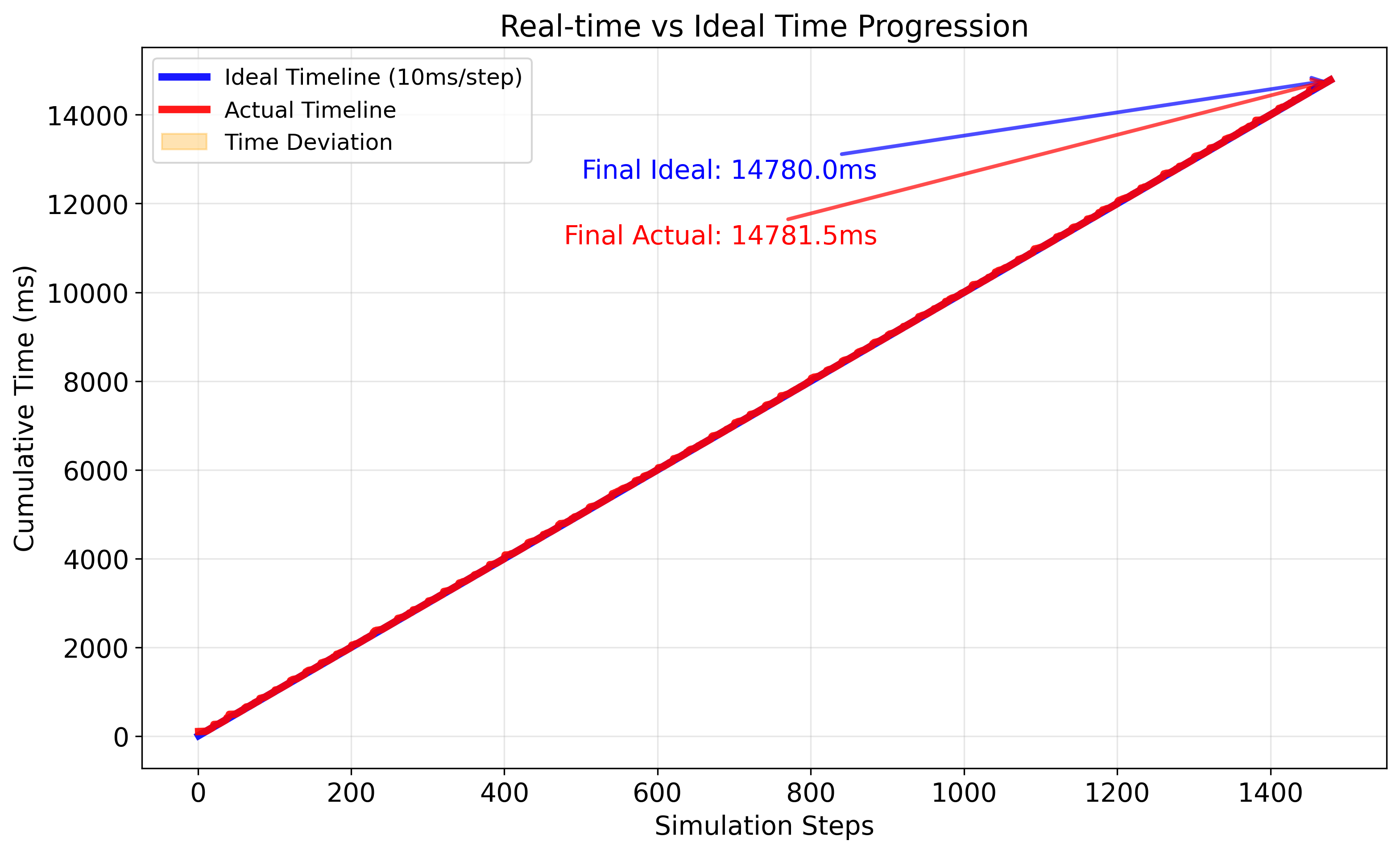}\label{fig:timing_progression_prop}}
    \hfill
    \subfloat[Naive: Timing Progression]{\includegraphics[width=0.48\linewidth]{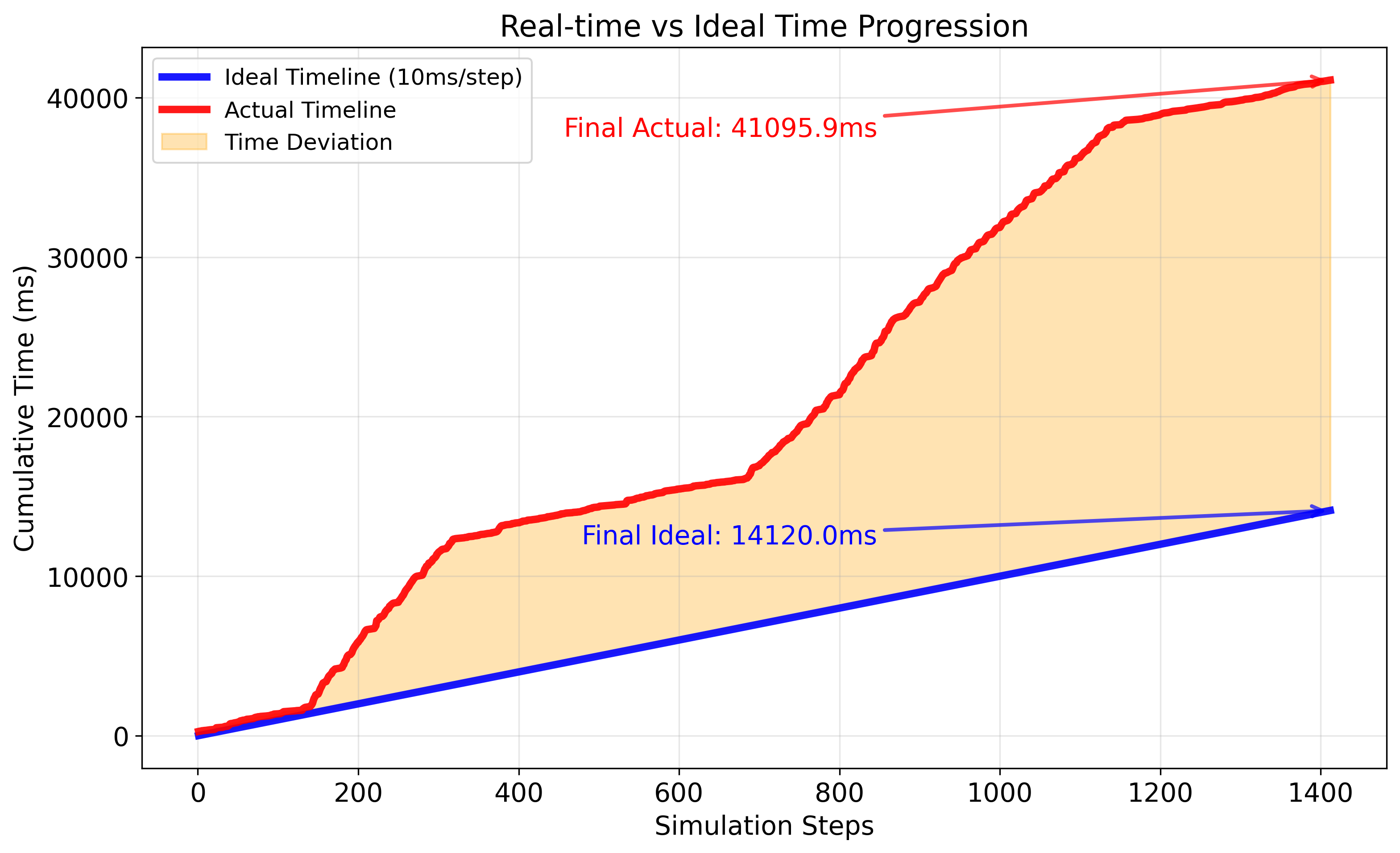}\label{fig:timing_progression_naive}}
    \\
    \subfloat[Proposed: Error Evolution]{\includegraphics[width=0.48\linewidth]{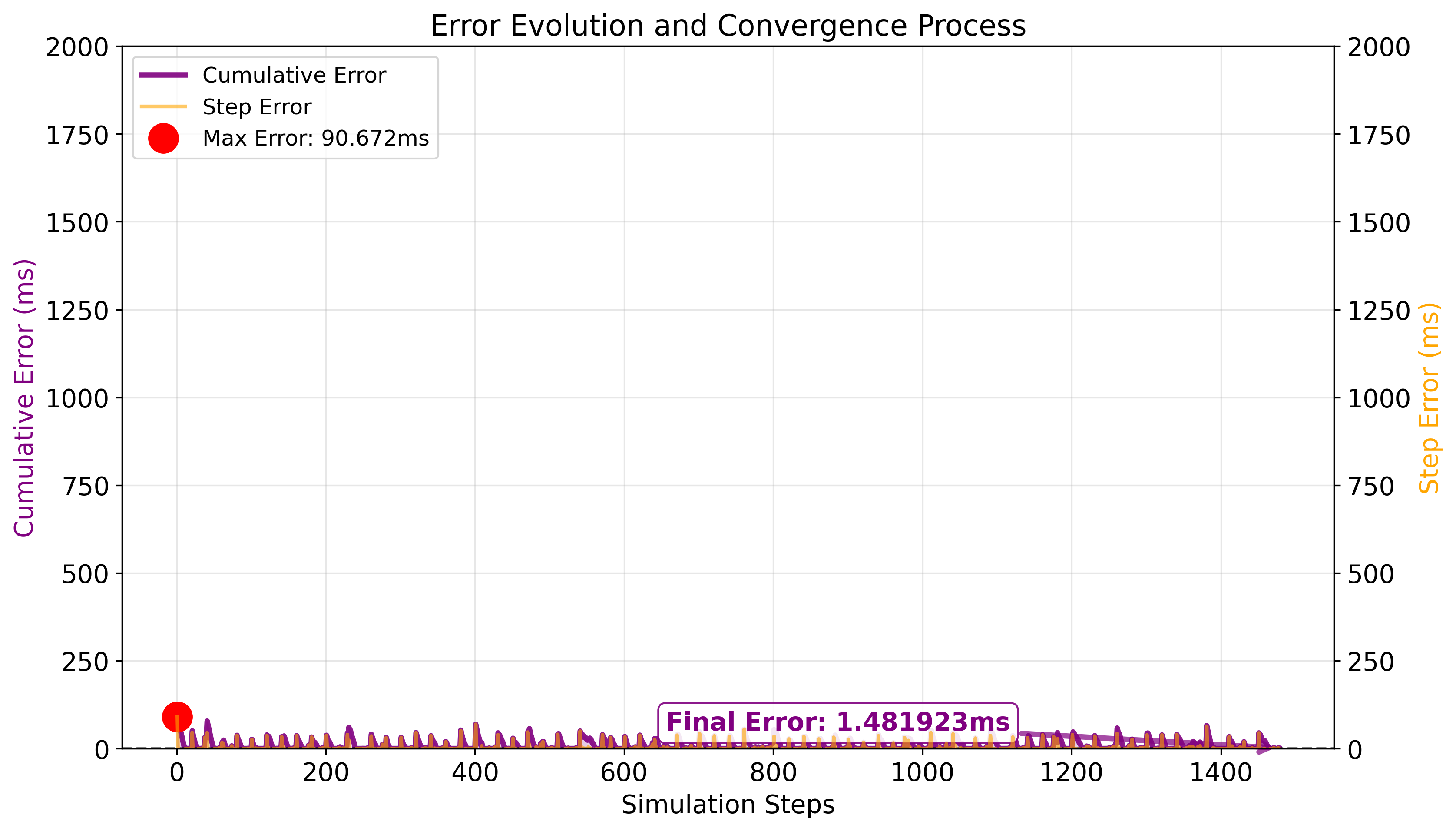}\label{fig:error_evolution_prop}}
    \hfill
    \subfloat[Naive: Error Evolution]{\includegraphics[width=0.48\linewidth]{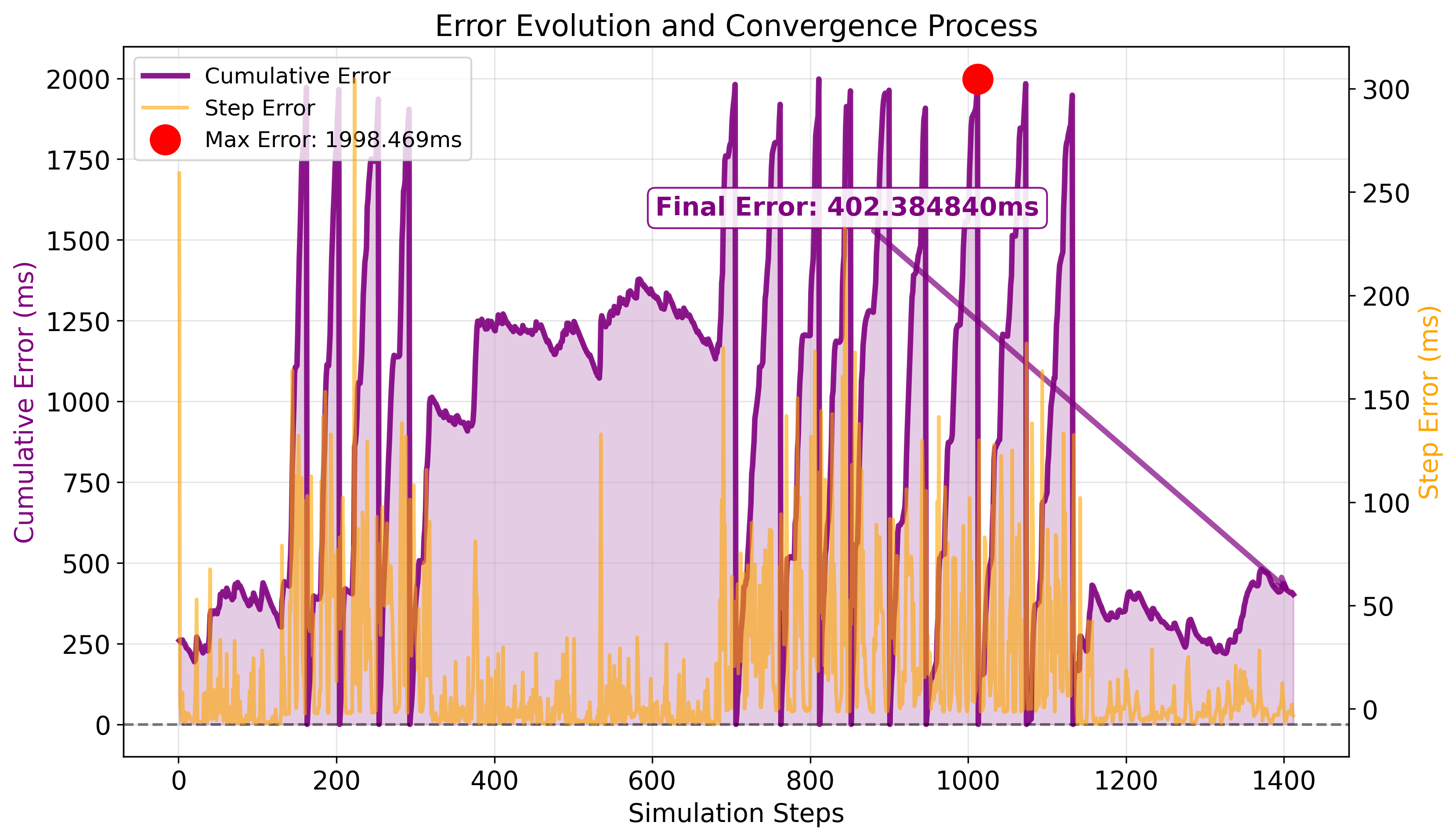}\label{fig:error_evolution_naive}}
    \caption{Comparison of timing synchronization performance. Left column: Proposed framework with synchronization mechanisms. Right column: Naive baseline without explicit synchronization. (a, b) Cumulative timeline comparison. (c, d) Evolution of timing errors.}
    \label{fig:timing_comparison}
\end{figure*}

To evaluate the temporal fidelity of the mechanisms described in Section~\ref{subsubsec:time_sync},
we compare the proposed framework against a naive baseline that lacks explicit synchronization control. The experiment is conducted in a single-AV scenario using the IDM planner. Fig.~\ref{fig:timing_comparison} visualizes the cumulative
ideal timeline and the realized wall-clock execution for both approaches, together
with the evolution of step-wise and accumulated timing errors. 
Table~\ref{tab:timing_comparison} summarizes the quantitative comparison. We define the
timing behavior metrics as follows:

\begin{itemize}
    \item \textbf{Final Error ($E_f$):} the absolute difference between simulation
    time and wall-clock time at the end of an experiment, indicating the residual
    drift after the synchronization mechanisms have acted.

    \item \textbf{Average Step Error ($\bar{e}_{\mathrm{step}}$):} the mean absolute
    deviation of the realized step duration from the target step size (e.g.,
    $10\,\mathrm{ms}$), reflecting the stability of loop pacing.

    \item \textbf{Maximum Cumulative Error ($E_{\max}$):} the maximum divergence
    between the ideal schedule and the realized execution over the full run,
    capturing the worst-case transient lag prior to recovery.

    \item \textbf{Timeout Ratio ($R_{\mathrm{timeout}}$):} the fraction of simulation
    steps for which computation exceeds the available time budget (target step size).
    Larger values indicate sustained overload relative to real-time capacity.

    \item \textbf{Real-Time Ratio ($R_{\mathrm{RT}}$):} the ratio between simulated
    time elapsed and wall-clock time elapsed. A value of $1.0$ indicates perfect
    real-time execution. We report the average ($\bar{R}_{\mathrm{RT}}$) and minimum
    ($R_{\mathrm{RT,min}}$) values to characterize overall performance and worst-case
    slowdowns.

    \item \textbf{Time Efficiency ($\eta_t$):} the ratio of the total simulated
    duration to the total wall-clock duration of the experiment, expressed as a
    percentage, providing a global measure of speed relative to real time.

    \item \textbf{Precision Ratio ($P_r$):} the ratio of the final timing error to
    the total wall-clock duration of the experiment, i.e.,
    $P_r = E_f / (\tau_K - \tau_0)$, which normalizes residual drift by the overall
    experiment length.
\end{itemize}

\begin{table}[t]
\caption{Quantitative Comparison of Timing Synchronization Performance}
\label{tab:timing_comparison}
\centering
\begin{tabular}{lcc}
\toprule
\textbf{Metric} & \textbf{Proposed} & \textbf{Naive} \\
\midrule
\multicolumn{3}{l}{\textit{Time Performance}} \\
Final Error ($E_f$) & 1.48 ms & 402.38 ms \\
Average Step Error ($\bar{e}_{step}$) & 4.10 ms & 22.84 ms \\
Maximum Cumulative Error ($E_{max}$) & 90.67 ms & 1998.47 ms \\
Timeout Ratio ($R_{timeout}$) & 37.7\% & 56.4\% \\
\midrule
\multicolumn{3}{l}{\textit{Real-time Performance}} \\
Final Real-Time Ratio ($R_{RT}$) & 1.00 & 0.97 \\
Average Real-Time Ratio ($\bar{R}_{RT}$) & 0.99 & 0.84 \\
Minimum Real-Time Ratio ($R_{RT,min}$) & 0.10 & 0.04 \\
\midrule
\multicolumn{3}{l}{\textit{Efficiency}} \\
Time Efficiency ($\eta_t$) & 99.99\% & 34.36\% \\
Precision Ratio ($P_r$) & $1.00 \times 10^{-4}$ & $2.85 \times 10^{-2}$ \\
\bottomrule
\end{tabular}
\end{table}

The results in Fig.~\ref{fig:timing_comparison} and Table~\ref{tab:timing_comparison} demonstrate the significant impact of the proposed synchronization mechanisms.
While the naive approach suffers from unbounded drift (reaching nearly 2000 ms) and poor time efficiency (34.36\%), the proposed framework effectively bounds the maximum cumulative error (approx. 200 ms) and maintains near-perfect time efficiency (99.99\%).
This confirms that the explicit synchronization strategies---wall-clock pacing, drift reset, and adaptive frame skipping---are essential for maintaining temporal fidelity in real-time human-in-the-loop simulations.

\raggedbottom
\section{Conclusion}

This paper presents CommonRoad-Game, a human-in-the-loop simulation framework tightly integrated with 
the CommonRoad ecosystem. Beyond serving as a testbed for 
autonomous driving motion planners, the framework enables systematic analysis of human driving
behaviors and supports the generation of interactive traffic scenarios within a unified and
reproducible setting.

In contrast to existing simulation platforms, CommonRoad-Game is, to the best of our knowledge,
the first framework to enable real-time human-in-the-loop interaction---through consumer input
devices such as a keyboard or a steering-wheel-and-pedals interface---directly within standardized
CommonRoad scenarios. This tight integration allows researchers to directly reuse standardized
scenarios, maps, and evaluation pipelines, thereby bridging the gap between offline
benchmarking and interactive, human-in-the-loop experimentation.

Compared to mature, large-scale autonomous driving simulators, the proposed
framework follows a lightweight design philosophy. By focusing on essential functionalities
required for real-time interaction and behavioral studies, it reduces system complexity
and computational overhead, making it more accessible for rapid prototyping and 
research-oriented use cases.

Furthermore, the framework explicitly addresses the often-overlooked discrepancy between
simulation time and wall-clock time. Through a multi-threaded architecture with explicit 
synchronization mechanisms, CommonRoad-Game ensures temporal fidelity during real-time operation, 
preventing unbounded time drift and maintaining timing errors within a bounded range. 
This design choice is critical for experiments involving human participants, 
where consistent real-time responsiveness is required.

\bibliographystyle{IEEEtran}
\bibliography{references}

\begin{thebibliography}{10}
\providecommand{\url}[1]{#1}
\csname url@samestyle\endcsname
\providecommand{\newblock}{\relax}
\providecommand{\bibinfo}[2]{#2}
\providecommand{\BIBentrySTDinterwordspacing}{\spaceskip=0pt\relax}
\providecommand{\BIBentryALTinterwordstretchfactor}{4}
\providecommand{\BIBentryALTinterwordspacing}{\spaceskip=\fontdimen2\font plus
\BIBentryALTinterwordstretchfactor\fontdimen3\font minus
  \fontdimen4\font\relax}
\providecommand{\BIBforeignlanguage}[2]{{%
\expandafter\ifx\csname l@#1\endcsname\relax
\typeout{** WARNING: IEEEtran.bst: No hyphenation pattern has been}%
\typeout{** loaded for the language `#1'. Using the pattern for}%
\typeout{** the default language instead.}%
\else
\language=\csname l@#1\endcsname
\fi
#2}}
\providecommand{\BIBdecl}{\relax}
\BIBdecl

\bibitem{schwarting2019social}
W.~Schwarting, A.~Pierson, J.~Alonso-Mora, S.~Karaman, and D.~Rus, ``Social
  behavior for autonomous vehicles,'' \emph{Proc. of the National Academy of
  Sciences}, vol. 116, no.~50, pp. 24\,972--24\,978, 2019.

\bibitem{althoff2017commonroad}
M.~Althoff, M.~Koschi, and S.~Manzinger, ``Common{R}oad: Composable benchmarks
  for motion planning on roads,'' in \emph{Proc. of the IEEE Intelligent
  Vehicles Symposium}, 2017, pp. 719--726.

\bibitem{dosovitskiy2017carla}
A.~Dosovitskiy, G.~Ros, F.~Codevilla, A.~Lopez, and V.~Koltun, ``{CARLA}: An
  open urban driving simulator,'' in \emph{Proc. of the Conference on Robot
  Learning}, 2017, pp. 1--16.

\bibitem{rong2020lgsvl}
G.~Rong, B.~H. Shin, H.~Tabatabaee, Q.~Lu, S.~Lemke, M.~Mo{\v{z}}eiko,
  E.~Boise, G.~Uhm, M.~Gerow, S.~Mehta \emph{et~al.}, ``{LGSVL} simulator: A
  high fidelity simulator for autonomous driving,'' in \emph{Proc. of the IEEE
  23rd International Conference on Intelligent Transportation Systems}, 2020,
  pp. 1--6.

\bibitem{shah2017airsim}
S.~Shah, D.~Dey, C.~Lovett, and A.~Kapoor, ``{AirSim}: High-fidelity visual and
  physical simulation for autonomous vehicles,'' in \emph{Field and Service
  Robotics}, 2018, pp. 621--635.

\bibitem{MAVLinkGuide}
{MAVLink Project}, ``{MAVLink} guide: Lightweight messaging protocol for
  drones,'' \url{https://mavlink.io/}, 2025, online.

\bibitem{li2023metadrive}
Q.~Li, Z.~Peng, L.~Feng, Q.~Zhang, Z.~Xue, and B.~Zhou, ``{MetaDrive}:
  Composing diverse driving scenarios for generalizable reinforcement
  learning,'' \emph{IEEE Transactions on Pattern Analysis and Machine
  Intelligence}, vol.~45, no.~3, pp. 3461--3475, 2023.

\bibitem{cao2020reinforcement}
Z.~Cao, E.~Biyik, W.~Z. Wang, A.~Raventos, A.~Gaidon, G.~Rosman, and D.~Sadigh,
  ``Reinforcement learning based control of imitative policies for
  near-accident driving,'' in \emph{Proc. of Robotics: Science and Systems},
  2020.

\bibitem{wuersching2024robust}
G.~W{\"u}rsching and M.~Althoff, ``Robust and efficient curvilinear coordinate
  transformation with guaranteed map coverage for motion planning,'' in
  \emph{Proc. of the IEEE Intelligent Vehicles Symposium}, 2024, pp.
  2694--2701.

\bibitem{treiber2000congested}
M.~Treiber, A.~Hennecke, and D.~Helbing, ``Congested traffic states in
  empirical observations and microscopic simulations,'' \emph{Physical Review
  E}, vol.~62, no.~2, pp. 1805--1824, 2000.

\bibitem{rajamani2012vehicle}
R.~Rajamani, \emph{Vehicle Dynamics and Control}, 2nd~ed.\hskip 1em plus 0.5em
  minus 0.4em\relax Boston, MA, USA: Springer, 2012.

\bibitem{gillespie1992fundamentals}
T.~D. Gillespie, \emph{Fundamentals of Vehicle Dynamics}.\hskip 1em plus 0.5em
  minus 0.4em\relax Warrendale, PA, USA: Society of Automotive Engineers, 1992.

\bibitem{pek2020commonroad}
C.~Pek, V.~Rusinov, S.~Manzinger, M.~C. {\"U}ste, and M.~Althoff,
  ``Common{R}oad drivability checker: Simplifying the development and
  validation of motion planning algorithms,'' in \emph{Proc. of the IEEE
  Intelligent Vehicles Symposium}, 2020, pp. 1013--1020.

\end{thebibliography}

\end{document}